\documentclass[10pt,twocolumn,letterpaper]{article}

\usepackage{iccv}
\usepackage{times}
\usepackage{epsfig}
\usepackage{graphicx}
\usepackage{amsmath}
\usepackage{amssymb}
\usepackage{listings}

\usepackage[pagebackref=true,breaklinks=true,letterpaper=true,colorlinks,bookmarks=false]{hyperref}

\iccvfinalcopy % *** Uncomment this line for the final submission

\def\iccvPaperID{3309} % *** Enter the ICCV Paper ID here

\ificcvfinal\fi

\begin{document}

%%%%%%%%% TITLE
\title{Implicit Integration of Superpixel Segmentation\\into Fully Convolutional Networks}

\author{Teppei Suzuki\\
Denso IT Laboratory, Inc.\\
Tokyo, Japan\\
{\tt\small tsuzuki@d-itlab.co.jp}
}

\maketitle
\ificcvfinal\thispagestyle{empty}\fi

%%%%%%%%% ABSTRACT
\begin{abstract}
Superpixels are a useful representation to reduce the complexity of image data.
However, to combine superpixels with convolutional neural networks (CNNs) in an end-to-end fashion, one requires extra models to generate superpixels and special operations such as graph convolution.
In this paper, we propose a way to implicitly integrate a superpixel scheme into CNNs, which makes it easy to use superpixels with CNNs in an end-to-end fashion.
% Our method preserves detailed information such as object boundaries, small and thin objects.
% A fully convolutional network (FCN) is a de facto standard architecture for dense prediction tasks.
% FCN-32, which is the simplest architecture, often misses detailed information such as object boundaries and small and thin objects due to downsampling operations; hence atrous convolution, also known as dilated convolution, or a trainable decoder is generally used to mitigate the information loss or recover lost resolution.
% As an alternative means of mitigating information loss, we implicitly integrate a superpixel segmentation scheme into FCNs.
Our proposed method hierarchically groups pixels at downsampling layers and generates superpixels.
Our method can be plugged into many existing architectures without a change in their feed-forward path because our method does not use superpixels in the feed-forward path but use them to recover the lost resolution instead of bilinear upsampling.
As a result, our method preserves detailed information such as object boundaries in the form of superpixels even when the model contains downsampling layers.
We evaluate our method on several tasks such as semantic segmentation, superpixel segmentation, and monocular depth estimation, and confirm that it speeds up modern architectures and/or improves their prediction accuracy in these tasks.
\end{abstract}

%%%%%%%%% BODY TEXT
\section{Introduction}
% Dense prediction tasks such as image segmentation, depth estimation, and image generation are fundamental tasks in the computer vision field.
Fully convolutional networks (FCNs) proposed by Long \textit{et al.}~\cite{fcn} have brought significant advances and become a reasonable choice for dense prediction tasks such as image segmentation, depth estimation.
Subsequently, various derivative models have been proposed~\cite{segnet,atrous,deeplab,deeplabv3,semflow,enet,dilated,pspnet} and applied to various dense prediction tasks.

Downsampling operation (\textit{e.g.}, max/average pooling and convolution with a stride of two or more) is applied to convolutional neural networks (CNNs) to reduce the resolution and efficiently expand the receptive field.
However, it loses detailed information such as the object boundaries and small, thin objects.
FCNs also use the downsampling layers and utilize a bilinear interpolation to recover the lost spatial resolution; hence, local structures are typically missed.
Many existing methods try to mitigate this local information loss by replacing the striding with atrous convolutions, also known as dilated convolutions~\cite{atrous,dilated}, or utilizing a trainable decoder to recover the lost resolution, which consists of the transposed convolution or the bilinear interpolation with some convolution layers~\cite{segnet,fpn,fcn,deconv,unet}.
These methods generate high-resolution segmentation maps but require high computational costs or additional layers compared with simple models such as FCN-32~\cite{fcn}.
\begin{figure}
    \centering
    \includegraphics[clip,width=1\hsize]{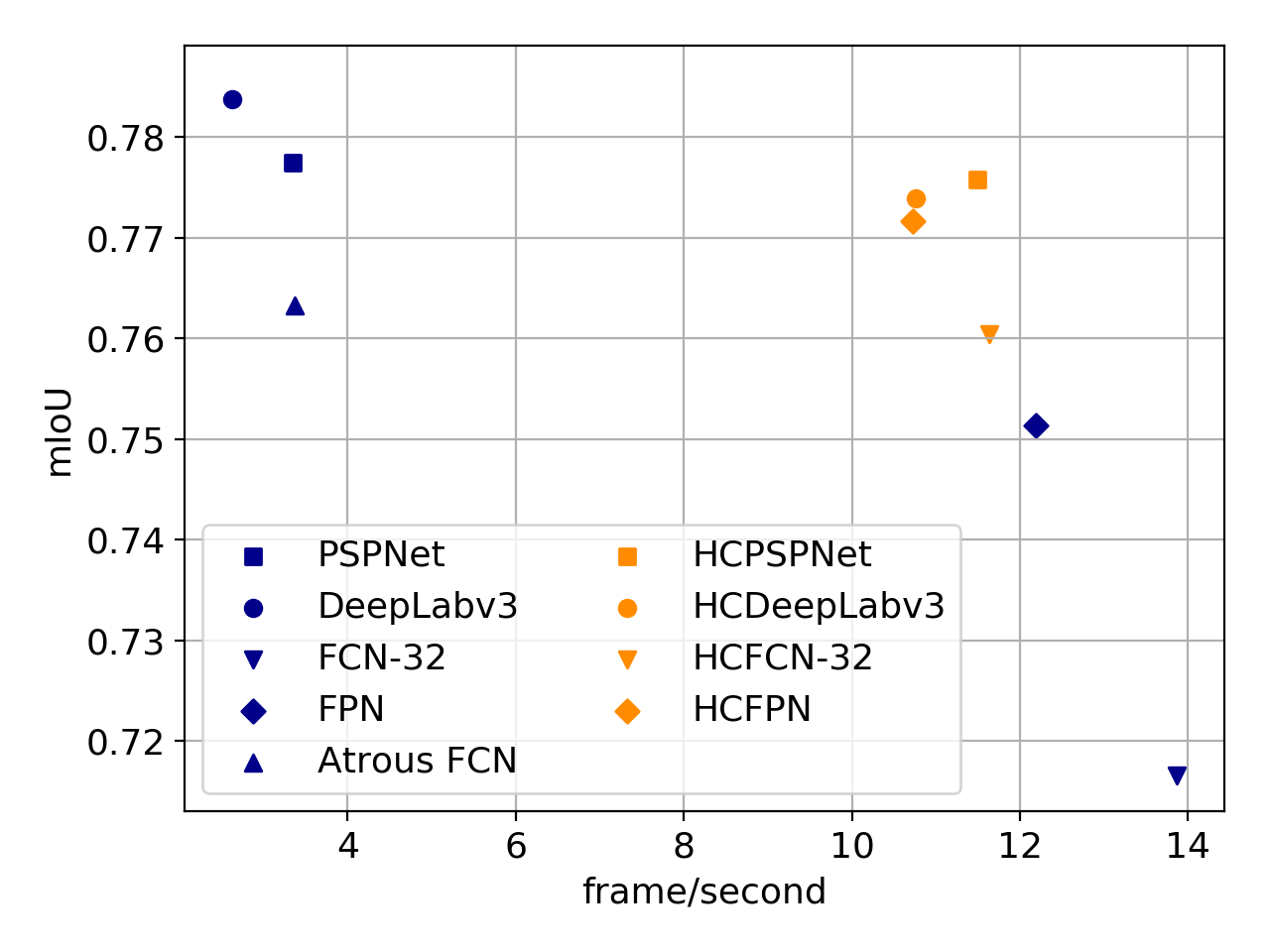}
    \caption{Inference speed versus mIoU on the Cityscapes~\cite{cityscapes} validation set with a 1024$\times$2048 input. All models use ResNet-101~\cite{resnet} as the backbone model. ``HC'' as a prefix denotes models using the proposed backbone. Our clustering module significantly reduces the inference time for PSPNet~\cite{pspnet} and DeepLabv3~\cite{deeplabv3}, and improves mIoU for FCN-32~\cite{fcn} and FPN~\cite{fpn}.}
    \label{fig:benchmark}
\end{figure}

% One of the inherent reason of information loss is that the model interprets the downsampled map as a regular grid image.
Classically, superpixels~\cite{spix_bench} are utilized as an efficient image representation, which can reduce the complexity of image data and preserve the local structure.
Some existing methods~\cite{spixpool,spixconv,spixfcn} combine superpixels with deep neural networks to reduce computational costs through the use of superpixel-based downsampling.
Due to the irregularity of superpixels, they require graph convolution~\cite{spix_conv2,monet,spixconv} to process the downsampled features.
Therefore, one needs to modify the feed-forward path of existing architectures and/or require an extra model to compute superpixels before downsampling.
% Moreover, some methods require the pre-computing of the superpixels by a classical superpixel segmentation method such as SLIC~\cite{slic}.
% Thus, existing CNN architectures must be modified to utilize superpixels or waste computational costs for calculation of them, although superpixels are efficient representation.

The aim of our work is to make it easy to use superpixels in CNNs.
% bridges the gap in combining superpixels with CNNs in an end-to-end fashion.
We interpret the general downsampling operation in CNNs as superpixel-based downsampling and generate superpixels by utilizing the sampled pixels as the seeds of superpixels.
As a result, our proposed method preserves the detailed information in the form of superpixels, and the superpixels are utilized for recovering the lost resolution.
The backbone architecture using the proposed method is more efficient than the backbone using atrous convolution, which is adopted by modern architectures such as PSPNet~\cite{pspnet} and Deeplabv3~\cite{deeplabv3}.
Unlike other methods combining superpixels with CNNs, our proposed method does not change the feed-forward path of base architectures because it does not explicitly use superpixels for downsampling.
Therefore, our proposed method can be plugged into many existing methods.
% We introduce a clustering procedure into the downsampling layers, which groups the pixels based on the sampled cluster centers, which mean the downsampled feature map.
% By integrating the clustering process in all downsampling layers, we implicitly build a superpixel segmentation scheme into CNNs.
%require specific operations or an additional model for superpixels.

% The advantages of our proposed method are three-fold: (1) it can be plugged into many existing dense prediction models without a change in their feed-forward path, (2) reduces computational costs for models, which replace striding with atrous convolution, without impairing the accuracy, and (3) improves the prediction accuracy of encoder-decoder architectures by integrating the proposed method and replacing bilinear upsampling in the decoder with our superpixel-based upsampling.
% \begin{itemize}
%     \item It can be plugged into many existing dense prediction models without a change in their feed-forward path,
%     \item It reduces computational costs for the models, which replace striding with atrous convolution, without impairing the accuracy.
%     \item It improves the prediction accuracy of encoder-decoder architectures by integrating the proposed method and replacing bilinear upsampling in the decoder with our cluster-based upsampling.
% \end{itemize}
The advantages of our proposed method are three-fold: (1) it can be plugged into many existing dense prediction models without a change in their feed-forward path, (2) reduce the computational costs for models using atrous convolution~\cite{pspnet,deeplabv3} by replacing their backbone with our proposed backbone, (3) improve the perdiction accuracy of encoder-decoder architecuters~\cite{fpn} by integrating the proposed method and replacing bilinear upsampling in the decoder with our superpixel-based upsampling.
In the case of the semantic segmentation task, our method speeds up or improves the mean intersection over union (mIoU) of modern architectures, as shown in Fig. \ref{fig:benchmark}.
Moreover, we confirm the effectiveness of our method in superpixel segmentation and monocular depth estimation in our experiments.

%-------------------------------------------------------------------------
\section{Related Work}
Long \textit{et al.} proposed fully convolutional networks (FCNs)~\cite{fcn} for semantic segmentation tasks, and other various derivative models have been proposed~\cite{atrous,deeplabv3,deeplabv3+,fpn,unet,dilated}.
FCNs are used for not only the segmentation tasks, but also various tasks such as depth estimation~\cite{dorn,monodepth,monodepth2}, optical flow estimation~\cite{flownet,flownet2,pwc}, superpixel segmentation~\cite{ssn,spixrim,seal,spixfcn}, and fundamental inverse problems (\textit{e.g.}, deblurring~\cite{deblurring2,deblurring} and inpainting~\cite{inpaint,inpaint2}).

FCN-32~\cite{fcn} is the simplest model, generating a prediction map with a reduced spatial resolution of 1/32.
Because FCN-32 uses the bilinear interpolation, which is a static and linear interpolation method, to recover the lost spatial resolution, local structures such as object boundaries are often missed.
This issue may be caused by the methods that generate the prediction map from coarse outputs by utilizing bilinear interpolation.
Therefore, to generate a high-resolution map, many existing methods utilize a trainable decoder or replace the striding with atrous convolution, also known as dilated convolution~\cite{atrous,dilated}.

The encoder-decoder models~\cite{segnet,fpn,deconv,unet} recover the lost resolution and local structure using a trainable decoder that consists of transposed convolutions or bilinear interpolations with some convolution layers.
The encoder-decoder models can generate a high-resolution map but require additional layers and parameters for the decoder.

Atrous convolution~\cite{atrous,dilated} is used to efficiently expand receptive fields instead of striding.
Many modern architectures, such as PSPNet~\cite{pspnet} and DeepLabv3~\cite{deeplabv3}, have only a few downsampling layers (typically three) and uses atrous convolution instead.
%atrous convolution instead of striding and generally generate a prediction map with an output stride of eight, meaning the resolution of the predicted map is eight times smaller than that of the input image.
Although such models avoid the information loss by removing downsampling layers and demonstrate effective results, they typically require high computational costs because they discard some downsampling operations and process a large number of pixels in intermediate layers.

As a detail-preserving complexity reduction for image data, superpixel segmentation is classically used, which groups pixels similar in color and other low-level properties.
Superpixels can preserve object boundaries and semantics, and some existing methods~\cite{spix_bilateral,spix_conv2,spixpool,spixconv,spixfcn,dgcnet} combine superpixels with deep neural networks.
Many of them explicitly use superpixels as a downsampling operation and need special operations such as graph convolution to process the downsampled image because of the irregularity of superpixels or require an additional model to compute superpixels before downsampling.

Our strategy also uses superpixels for reducing the complexity of image data and preserving the local structure.
Unlike the many existing methods utilizing superpixels, our proposed method does not change the feed-forward path and does not require an additional model because we do not explicitly use superpixels for downsampling.
% Our method views a general downsampling operation as the superpixel-based downsampling and generates superpixels that are not used for downsampling but for recovering the lost resolution.
% As a result, our method gives FCNs the detail-preserving property, and demonstrates comparable accuracy to the method using atrous convolution based backbones even when the five downsampling layers are involved.

\textit{In principle}, as logn as the upsampling modules such as bilinear interpolation and transposed convolution are used, our method can be incorporated with existing models and modules (\textit{e.g.}, preserving or recovering detailed information~\cite{sfnet,gun,igun,fullres,takikawa} and global context aggregation~\cite{deeplabv3,nonlocal,dgcnet,pspnet}) by replacing the backbone and basic upsampling modules with our proposed backbone and superpixel-based upsampling.
In this paper, we verify the effectiveness of our method for the simple and widely used models and modules~\cite{fcn,fpn,pspnet,deeplabv3,ssn,dorn}.

\begin{figure*}
    \centering
    \includegraphics[clip, width=1\hsize]{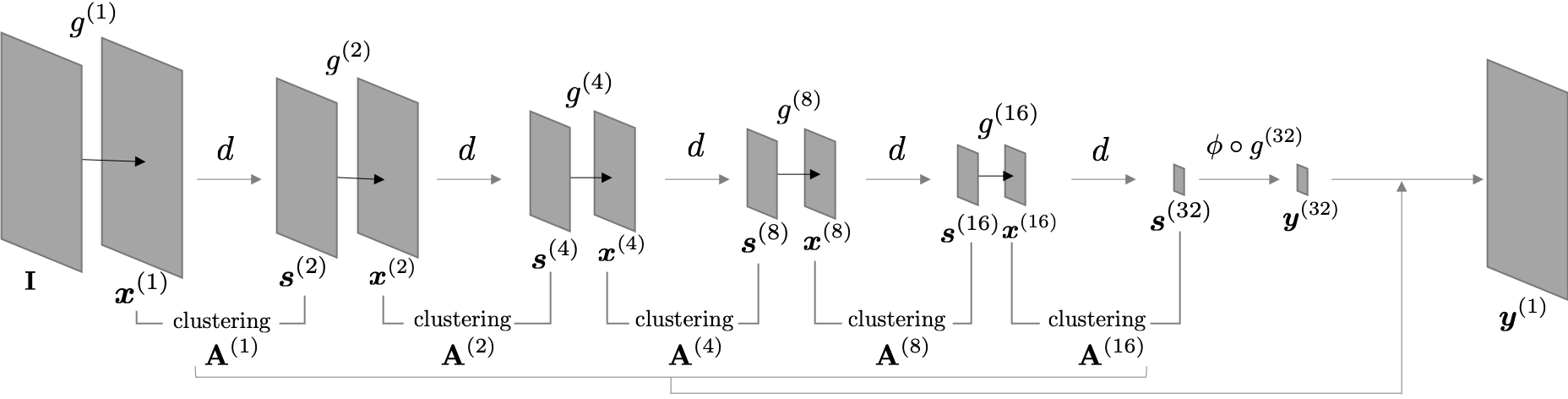}
    \caption{The proposed backbone using our hierarchical clustering. The proposed method groups pixels at downsampling layers, and forms a set of pixels as assignment matrices. The model predicts target values for a set of pixels. Unlike existing methods combining superpixel segmentation and neural networks, our method does not use superpixels for downsampling, explicitly. Therefore, our method can be plugged into existing architectures without a change in their feed-forward path. Superpixels are only used to recover the resolution by Eq. \eqref{eq:decode} instead of bilinear interpolation.}
    \label{fig:overview}
\end{figure*}

%-------------------------------------------------------------------------
\section{Method}
Our motivation is to mitigate the information loss by utilizing the superpixel property.
We show the overview of our proposed method in Fig. \ref{fig:overview}.
Our approach views the general downsampling as superpixel-based downsampling and groups pixels by utilizing the sampled pixels as superpixel seeds.
The coarse prediction map is decoded to a fine resolution based on the generated superpixels.

By integrating our proposed method into downsampling layers, the model generates superpixels hierarchically and predicts the target values for superpixels.
Our method does not change the feed-forward path of a base architecture because superpixels are not used in the feed-forward path, but the superpixels are only used for recovering the lost resolution instead of the bilinear interpolation.
Because of the property of superpixels, the model can preserve the detailed information.
% preserve the object boundaries and small and thin objects, the model can recover the detailed information.

\subsection{Preliminary}
Let $\mathbf{I}\in\mathbb{R}^{HW\times 3}$ be an RGB image where $H$ and $W$ denote the image height and width and $\boldsymbol{x}^{(s)}_i\in\mathbb{R}^N$ be a $N$-dimension feature vector of an $i$-th pixel in the feature map, where $s$ denotes an output stride, namely, the resolution of $\boldsymbol{x}^{(s)}$ is $(H/s, W/s)$.
FCNs consist of the blocks built on convolution layers, ReLU activation and downsampling layers such as max-pooling and convolution layers with a stride of two or more.
Let $g^{(s)}$ and $d^{(s)}:\mathbb{R}^{\frac{HW}{s^2}\times N}\rightarrow\mathbb{R}^{\frac{HW}{s'^2}\times M},\ s<s'$ be the Conv+ReLU blocks and a downsampling layer that reduces the spatial resolution.
For simplicity, we define the downsampling layer as the operation reducing the resolution in half, from $(\frac{H}{s},\frac{W}{s})$ to $(\frac{H}{2s}, \frac{W}{2s})$, without loss of generality, and we describe the downsampling layer as just $d(\cdot)$ in the rest of this paper.
Note that the feature dimension of the downsampled feature map may change when using the strided convolution as the downsampling.
Then, as shown in Fig. \ref{fig:overview}, the prediction map generated by FCN-32~\cite{fcn} is defined as $\boldsymbol{y}^{(32)}=\phi\circ g^{(32)}\circ d\circ\cdots \circ g^{(2)}\circ d\circ  g^{(1)}(\mathbf{I})$, where $\phi(\cdot)$ is the mapping into target values.
Our method does not change the feed-forward path, meaning the mapping from $\mathbf{I}$ to $\boldsymbol{y}^{(32)}$.

\subsection{Clustering Procedure}
Our strategy is to group pixels at downsampling layers, predict the target values for clusters, and share the predicted value with pixels belonging to the corresponding cluster.
The $j$-th cluster at the output stride of $s$ is defined as $c_j^{(s)}=\{i|\forall k, \mathcal{S}(\mathbf{W}^{(s)}\boldsymbol{x}^{(s)}_i,\tilde{\mathbf{W}}^{(2s)}\boldsymbol{s}^{(2s)}_j)\geq \mathcal{S}( \mathbf{W}^{(s)}\boldsymbol{x}^{(s)}_i,\tilde{\mathbf{W}}^{(2s)}\boldsymbol{s}^{(2s)}_k)\}$, where $\boldsymbol{s}^{(2s)}_j$ denotes the $j$-th pixel of the downsampled feature map, $\boldsymbol{s}^{(2s)}=d(\boldsymbol{x}^{(s)})$, and $\mathcal{S}$ denotes a similarity function that is defined as the cosine similarity in our experiments.
$\mathbf{W}^{(s)}\in\mathbb{R}^{K\times N}$ and $\tilde{\mathbf{W}}^{(2s)}\in\mathbb{R}^{K\times M}$ are learnable weight matrices for $\boldsymbol{x}^{(s)}\in\mathbb{R}^{\frac{HW}{s^2}\times N}$ and $\boldsymbol{s}^{(2s)}\in\mathbb{R}^{\frac{HW}{4s^2}\times M}$, which map feature vectors into a $K$-dimension space.
$K$ is set to 64 for our experiments.
This clustering process is executed in downsampling layers, and generates the set of pixels in each resolution, $\{c^{(s)}\}$.

Although the cluster to which an $i$-th pixel belongs can be obtained by $\arg\max_k\mathcal{S}(\mathbf{W}^{(s)}\boldsymbol{x}^{(s)}_i, \tilde{\mathbf{W}}^{(2s)}\boldsymbol{s}^{(2s)}_k)$, it is non-differentiable.
Therefore, to train the model in an end-to-end manner, we relax it by the widely used technique of the temperature softmax function~\cite{gsoftmax,conc}.
We first formulate hard clustering and next describe its relaxed version.

The hard clustering problem is defined as follows:
\begin{align}
    \label{eq:hard_cls}
    \mathbf{A}^{(s)\ast}=\underset{\mathbf{A}^{(s)}\in\{0,1\}^{U\times V}}{\arg\max} \sum_{ij}\mathbf{A}^{(s)}_{ij}\mathbf{S}^{(s)}_{ij},\ s.t.,\ \sum_j\mathbf{A}^{(s)}_{ij}=1,
\end{align}
where $\mathbf{A}^{(s)\ast}$ and $\mathbf{S}^{(s)}\in\mathbb{R}^{U\times V}$ denote an assignment matrix and a similarity matrix that is defined as $\mathbf{S}^{(s)}_{ij}=\mathcal{S}(\mathbf{W}^{(s)}\boldsymbol{x}^{(s)}_i,\tilde{\mathbf{W}}^{(2s)}\boldsymbol{s}^{(2s)}_j)$.
$U$ and $V$ denote the number of pixels in $\boldsymbol{x}^{(s)}$ and $\boldsymbol{s}^{(2s)}$, respectively.
Then, we can decode the downsampled feature map to the fine resolution feature map based on the clusters, as follows:
\begin{align}
    \label{eq:single_decode}
    \tilde{\boldsymbol{x}}^{(s)}=\mathbf{A}^{(s)\ast}\boldsymbol{x}^{(2s)}.
\end{align}
The model predicts target values from the coarsest feature map, $\boldsymbol{y}^{(32)}=\phi(\boldsymbol{x}^{(32)})$, and then, the coarsest prediction can be decoded to the original resolution by recursively decoding $\boldsymbol{y}^{(32)}$ as follows:
\begin{align}
    \label{eq:decode}
    \boldsymbol{y}^{(1)} = \prod_{s'=\{16,8,4,2,1\}}\mathbf{A}^{(s')\ast}\boldsymbol{y}^{(32)}.
\end{align}
Note that we plug the clustering modules into a part of all the downsampling layers in our experiments.
Therefore, we decode the prediction with the bilinear upsampling to the original resolution after decoding it to the plausible fine resolution by Eq. \eqref{eq:decode}.

Unfortunately, the clustering  procedure is non-differentiable because of $\arg\max$ in Eq. \eqref{eq:hard_cls}.
Thus, we relax the assignment matrix $\mathbf{A}^{(s)\ast}$ to a soft assignment matrix $\tilde{\mathbf{A}}^{(s)}\in(0, 1)^{U\times V}$.
We define the soft assignment from $\boldsymbol{x}^{(s)}_i$ to the sampled cluster seed $\boldsymbol{s}^{(2s)}_j$ as follows:
\begin{align}
    \tilde{\mathbf{A}}^{(s)}_{ij}=\frac{\exp\left(\mathcal{S}\left( \mathbf{W}^{(s)}\boldsymbol{x}^{(s)}_i, \tilde{\mathbf{W}}^{(2s)}\boldsymbol{s}^{(2s)}_j \right)/\tau \right)}{\sum_k\exp\left(\mathcal{S}\left( \mathbf{W}^{(s)}\boldsymbol{x}^{(s)}_i, \tilde{\mathbf{W}}^{(2s)}\boldsymbol{s}^{(2s)}_k \right)/\tau \right)},
\end{align}
where $\tau$ is a temperature parameter.
If $\tau\rightarrow 0$, $\tilde{\mathbf{A}}^{(s)}$is equal to $\mathbf{A}^{(s)\ast}$.
We set $\tau$ to 0.07 for our experiments.
The dense prediction map is generated by Eq. \eqref{eq:decode} using $\tilde{\mathbf{A}}^{(s)}$ instead of $\mathbf{A}^{(s)\ast}$.
When calculating the loss between ground-truth labels and the prediction map generated by Eq. \eqref{eq:decode} with $\tilde{\mathbf{A}}^{(s)}$, the loss is fully backpropable, and the model can be trained in an end-to-end manner.

\begin{figure*}
    \centering
    \begin{tabular}{c@{\hspace{1mm}}c@{\hspace{1mm}}c@{\hspace{1mm}}c@{\hspace{1mm}}}
        \includegraphics[clip,width=0.245\hsize]{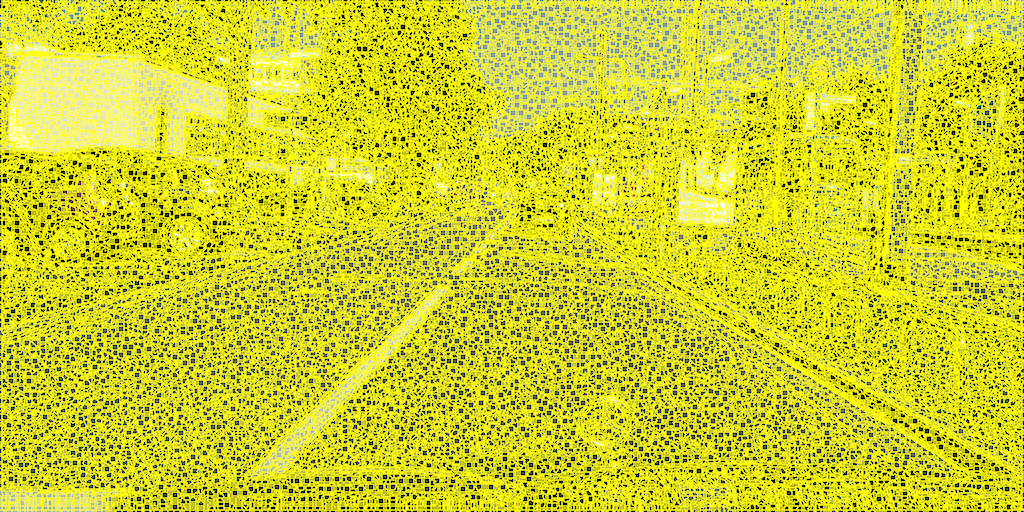} &
        \includegraphics[clip,width=0.245\hsize]{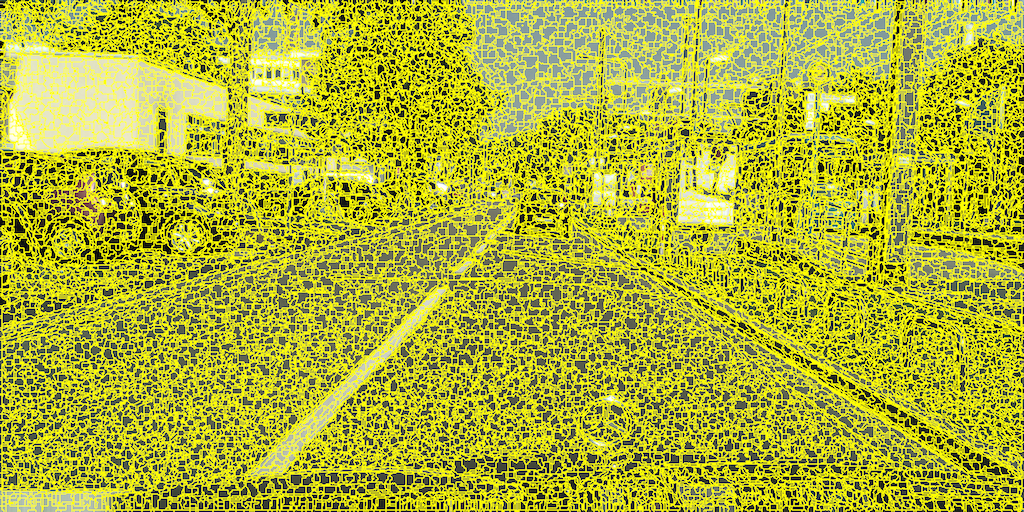} &
        \includegraphics[clip,width=0.245\hsize]{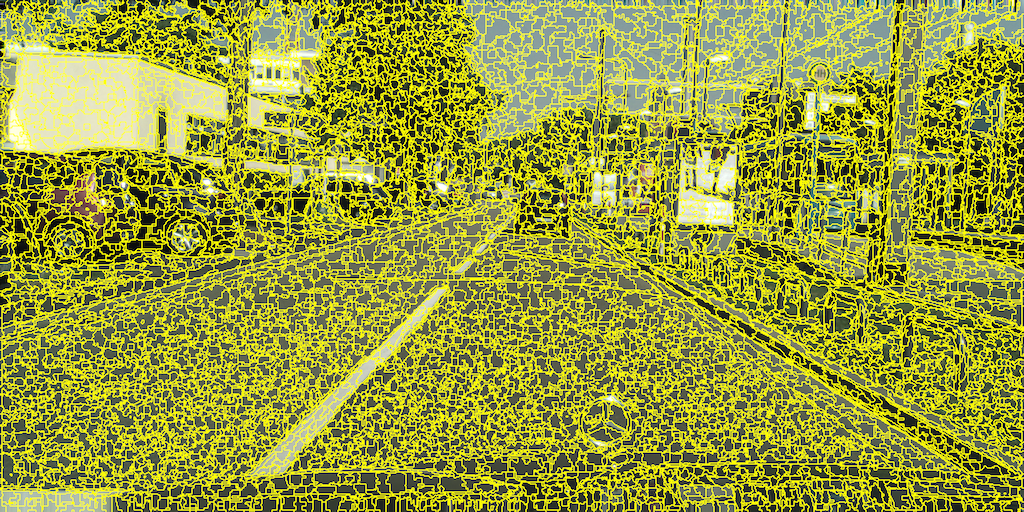} &
        \includegraphics[clip,width=0.245\hsize]{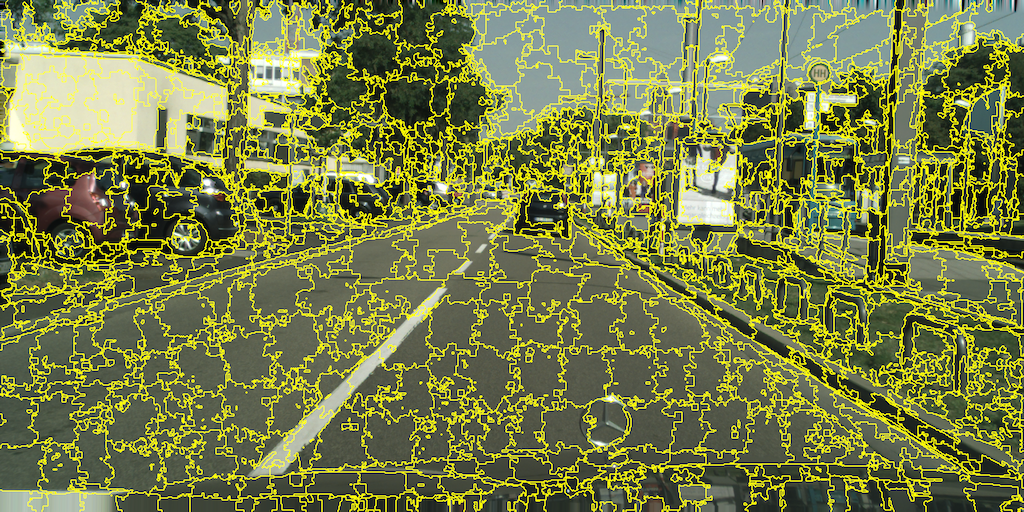}\\
        \includegraphics[clip,width=0.245\hsize]{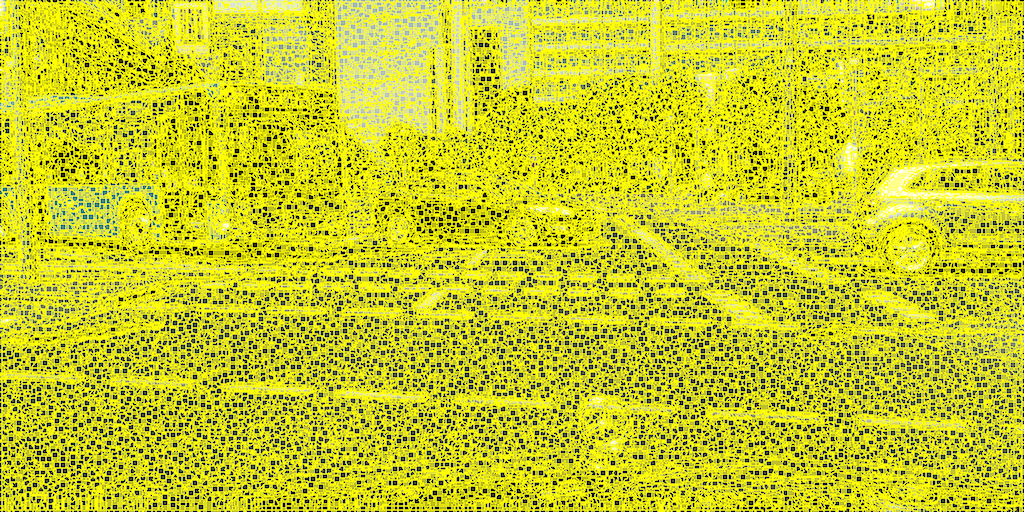} &
        \includegraphics[clip,width=0.245\hsize]{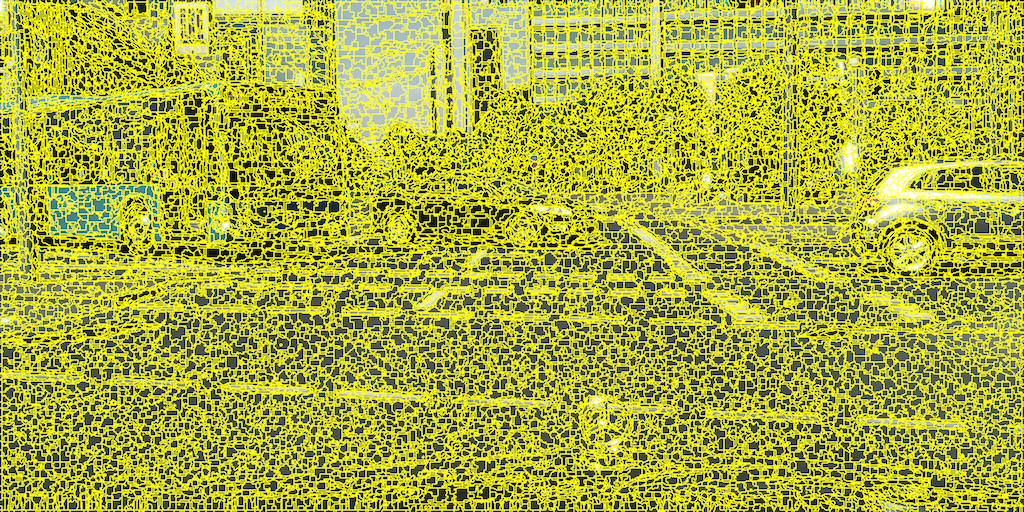} &
        \includegraphics[clip,width=0.245\hsize]{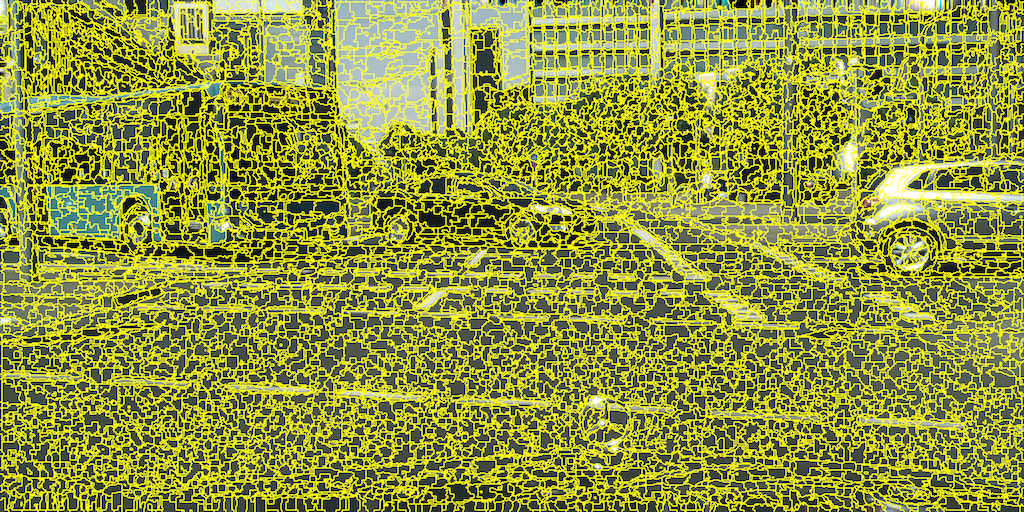} &
        \includegraphics[clip,width=0.245\hsize]{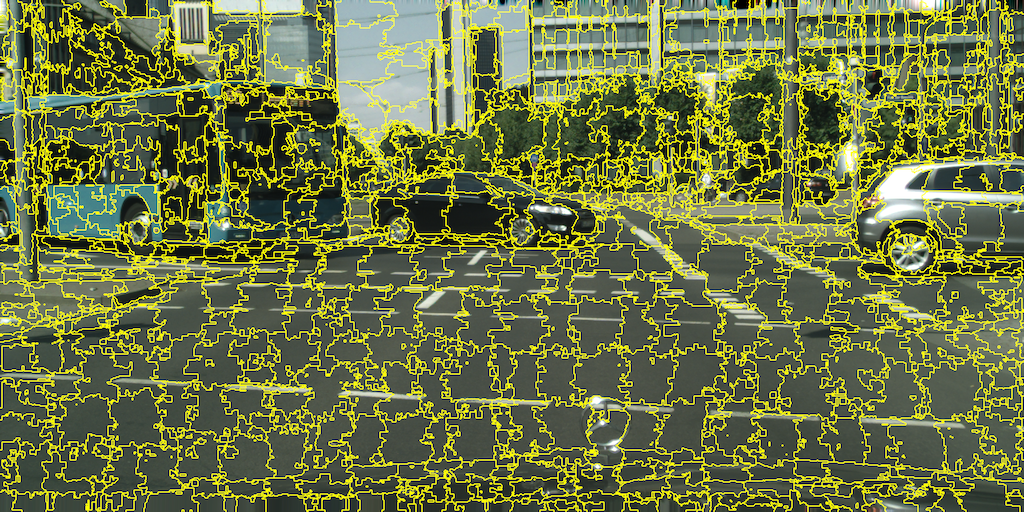}\\
        \includegraphics[clip,width=0.245\hsize]{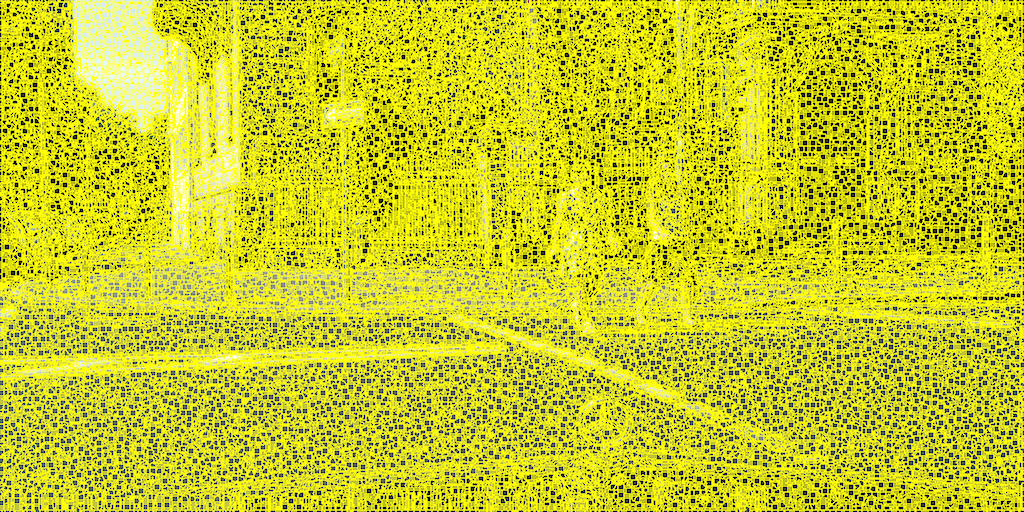} &
        \includegraphics[clip,width=0.245\hsize]{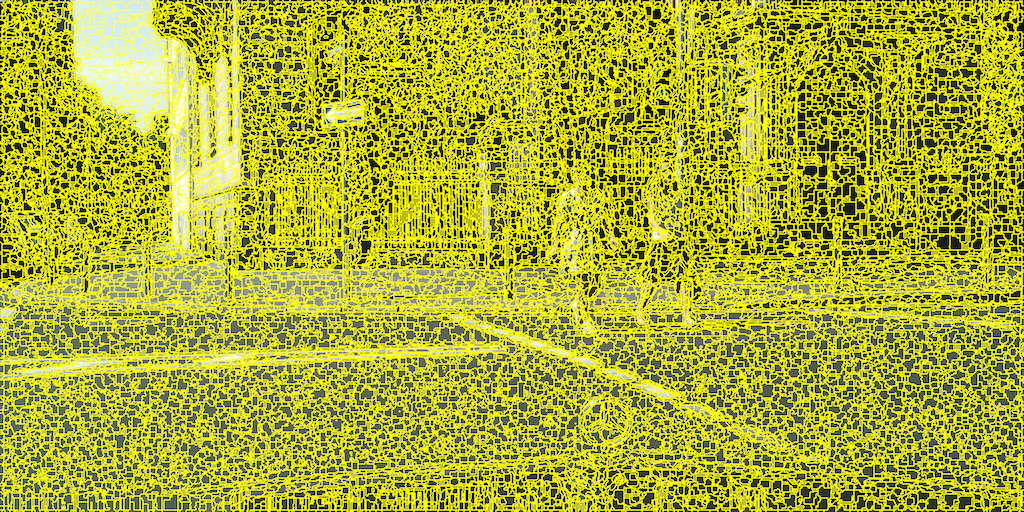} &
        \includegraphics[clip,width=0.245\hsize]{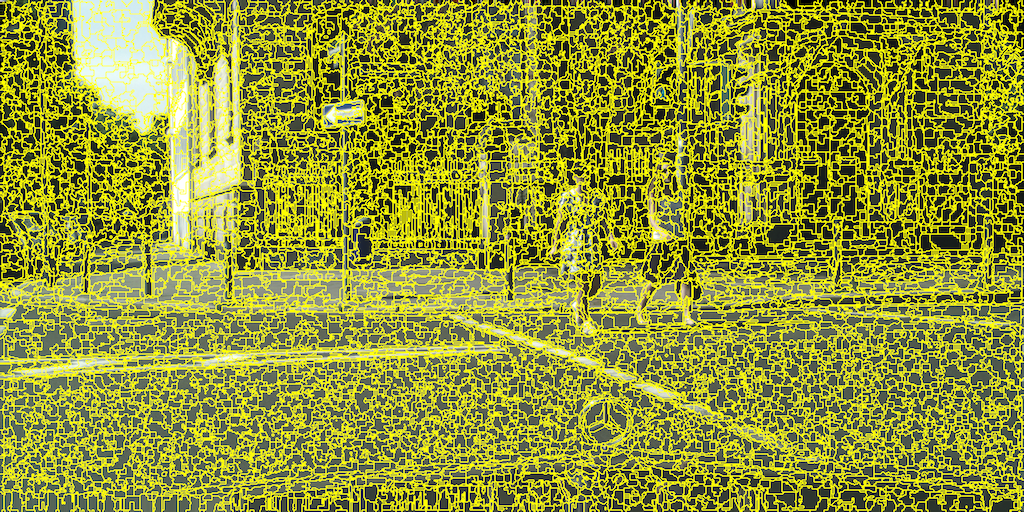} &
        \includegraphics[clip,width=0.245\hsize]{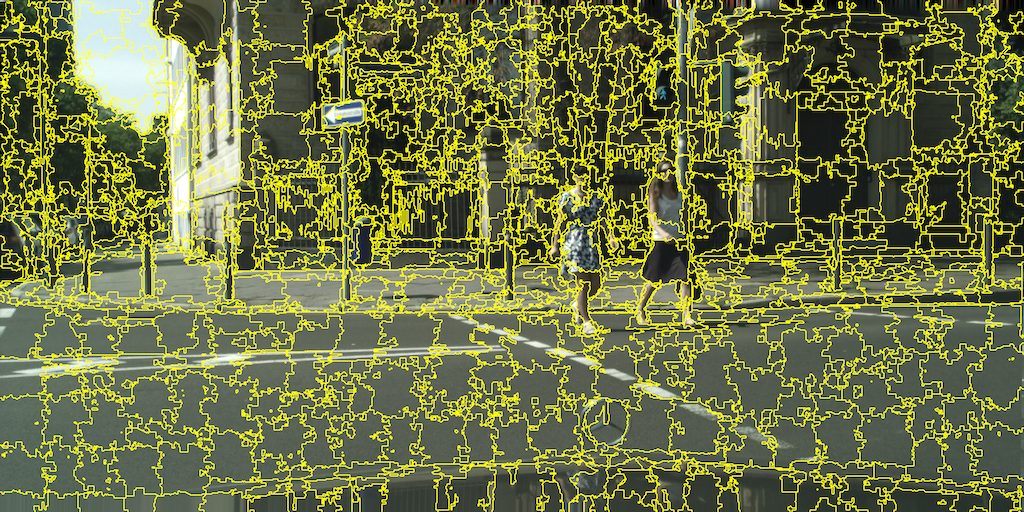}\\
        $c^{(2)}$ & $c^{(4)}$ & $c^{(8)}$ & $c^{(16)}$
    \end{tabular}
    \caption{An example result of hierarchical clustering. We visualize cluster boundaries as yellow lines. These clusters are generated by FCN-32 with the ResNet-101 backbone combined with our proposed method.}
    \label{fig:clustering_res}
\end{figure*}

We visualize hierarchical clustering in Fig. \ref{fig:clustering_res}.
The clusters are generated by FCN-32 with our proposed method trained on Cityscapes~\cite{cityscapes}.
Pixels are grouped hierarchically, and the model forms $\frac{HW}{32^2}$ clusters in the end.
The clusters preserve object boundaries and small and thin objects, such as signs and poles.

\subsection{Practical Issues and Their Solutions}
\begin{figure}
    \centering
    \includegraphics[clip,width=1\hsize]{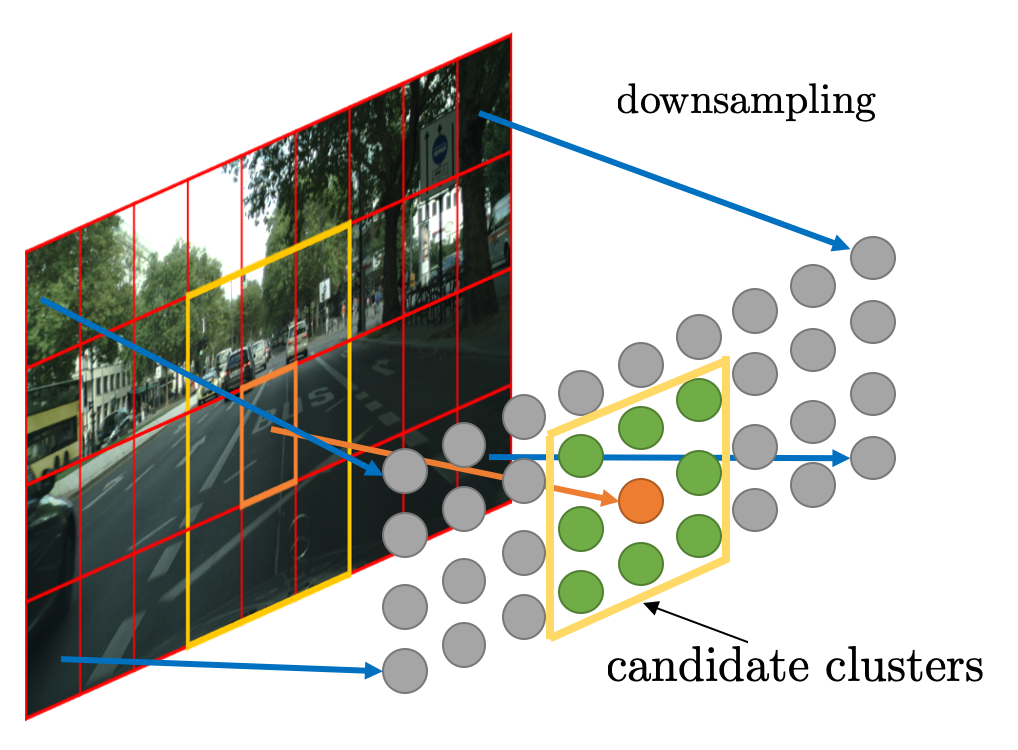}
    \caption{Illustration of the candidate clusters. Grid is defined as stride, meaning that if stride is two, then each cell has $2\times 2$ pixels. Circles indicate the cluster seeds sampled from corresponding regions. The candidate clusters to which pixels in the orange cell belong are defined as the orange circle and eight neighbors.}
    \label{fig:candidate}
\end{figure}

The soft assignment matrix is not computationally applicable; for example, when the resolution of an input image is $512\times 512$ and downsampled to $256\times 256$, the soft assignment matrix requires 64GB in the single-precision floating-point number.
Therefore, we restrict the number of candidate clusters to only nine surrounding clusters, as shown in Fig. \ref{fig:candidate}, and bring down the size of $\tilde{\mathbf{A}}^{(s)}$ from $\frac{HW}{s^2}\times  \frac{HW}{4s^{2}}$ to $\frac{HW}{s^2}\times9$ elements.
This technique is also used in existing superpixel segmentation methods~\cite{slic,ssn,spixfcn}.

The flexibility of the downsampling operation is important for the clustering since the downsampled pixels are used as the cluster seeds.
However, static downsampling operations such as max-pooling and strided convolution may not sample effective seeds for clustering because they simply sample pixels from a fixed region.
Therefore, we use the modulated deformable convolution (DCNv2)~\cite{dconv2} with a stride of two as the downsampling operation, which can adaptively change kernel shapes and weights by the learnable offsets and modulation parameter generator.
We verify the effectiveness of DCNv2 for our proposed method in Appendix.
Note that the use of DCNv2 is not necessary but optional, and an ordinary convolution layer works well; in fact, the ordinary convolution with our method achieves better accuracy than the model using the atrous convolution.

%-------------------------------------------------------------------------
\section{Experiments}
We first evaluate our method on the Cityscapes~\cite{cityscapes} validation set with various settings as an ablation study, and we next integrate our method into modern architectures and compare them on semantic segmentation, superpixel segmentation, and depth estimation tasks.
For a fair comparison, we have reproduced all baseline models in PyTorch~\cite{pytorch}.
We use ``HC'' as a prefix to indicate the model using the proposed backbone illustrated in Fig. \ref{fig:overview}.
Moreover, we use OS to denote the output stride, meaning that the ratio of the spatial resolution of the input image to the resolution of the raw prediction map.

\subsection{Ablation Study}
\label{sec:exp}
In the ablation study, we validate whether our method has the detail-preserving property like atrous convolution without downsampling.
We train models with fine training data in Cityscapes to minimize the cross-entropy loss.
We set the initial learning rate to 0.01 and decay it with ``poly'' learning rate policy where the initial learning rate is multiplied by $(1-\frac{iter}{max\_iter})^{0.9}$.
We employ momentum SGD as optimizer and set the momentum to 0.9.
We train models for 50K iterations with a batch size of 16.
We use random resize between 0.5 and 2.0, random crop with a crop size of 1024$\times$1024, and random horizontal flip as the data augmentation.
Moreover, we use the same auxiliary loss as PSPNet~\cite{pspnet}.

We analyze our soft clustering module plugged into FCN-32.
We use ResNet-18 and ResNet-50 as the backbone, because ResNet~\cite{resnet} is widely used for many existing models as a backbone architecture, and the difference between ResNet-18 and ResNet-50 is not only the number of the building blocks but also the structure of them.
As an evaluation metric, we use the mean intersection over union (mIoU), which is a standard metric to quantify the semantic segmentation performance.

We first evaluate the soft clustering module with various sampling branches.
Our module utilizes downsampled feature maps, but the downsampling operations in ResNet are embedded into the building blocks, and we have some choices for obtaining the downsampled feature maps.
The building blocks including downsampling are defined as $B(\boldsymbol{x})=I(\boldsymbol{x})+R(\boldsymbol{x})$, where $I(\cdot)$ denotes a linear projection called identity mapping and $R(\cdot)$ denotes residual mapping.
We compare $I(\boldsymbol{x})$, $R(\boldsymbol{x})$, and $B(\boldsymbol{x})$ as downsampling branches.
We show the results in Tab. \ref{tab:various_branch}.
Note that the soft clustering modules are plugged into conv4\_x and conv5\_x in ResNet.
The feature map obtained from identity mapping is slightly worse than the others.
The residual mapping and the block are superior or inferior depending on the architecture, but the difference between them is less than when comparing them to the identity mapping.
We use the block as the sampling branch in the remainder of the experiment.
\begin{table}[t]
    \centering
    \begin{tabular}{c|c|c}
        sampling branch & backbone & mIoU \\ \hline
        identity & ResNet-18 & 71.21 \\
        & ResNet-50 & 72.26\\ \hline
        residual & ResNet-18 & 72.43 \\
        & ResNet-50 & \textbf{73.20}\\ \hline
        block & ResNet-18 & \textbf{72.70}\\
        & ResNet-50 & 72.93\\
    \end{tabular}
    \caption{Results on various downsampling branches.}
    \label{tab:various_branch}
\end{table}
\begin{figure}
    \centering
    \includegraphics[clip,width=1\hsize]{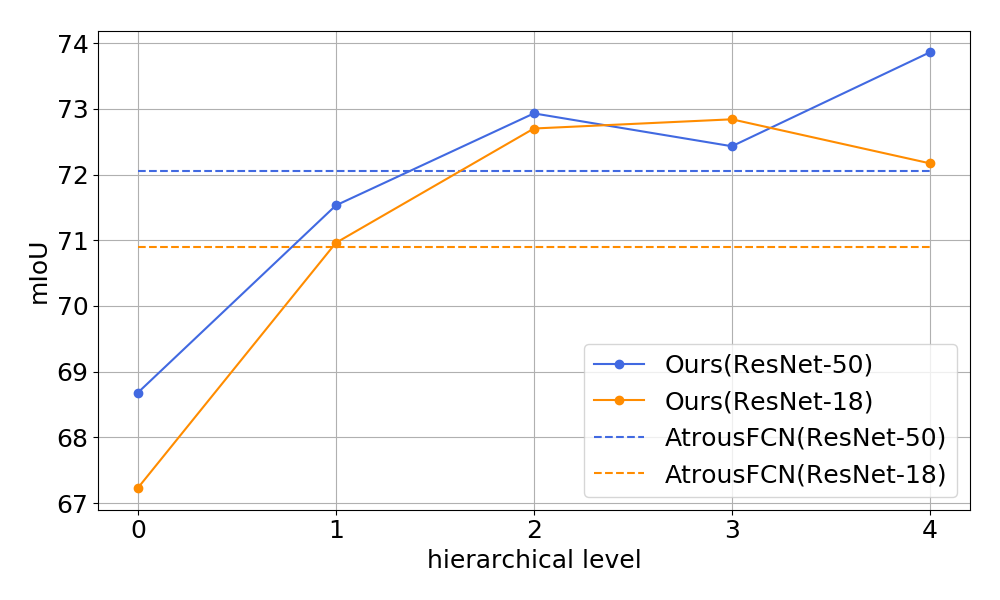}
    \caption{mIoU for various hierarchical levels for ResNet-18 and ResNet-50 as the backbone. The model corresponds to FCN-32~\cite{fcn} when the level is zero. Our proposed method with a level of two outperforms AtrousFCN for both backbones.}
    \label{fig:level}
\end{figure}

We then evaluate mIoU for the proposed method with various hierarchical levels.
We increase the number of the proposed modules from conv5\_x to conv2\_x in the backbone ResNet of FCN-32 and report their mIoU.
We show the results in Fig. \ref{fig:level}.
AtrousFCN is a model replacing the striding of conv4\_x and conv5\_x in the backbone of FCN-32 with atrous convolution.
Similar to ours, AtrousFCN preserves the detailed information without a change in the feed-forward path.
Our proposed method with one or two clustering modules outperforms AtrousFCN and further improves mIoU by increasing the hierarchical level.
Moreover, our method with ResNet-18 shows higher mIoU than AtrousFCN with ResNet-50 when the hierarchical level is two or more.

\begin{figure}
    \centering
    \includegraphics[clip,width=1\hsize]{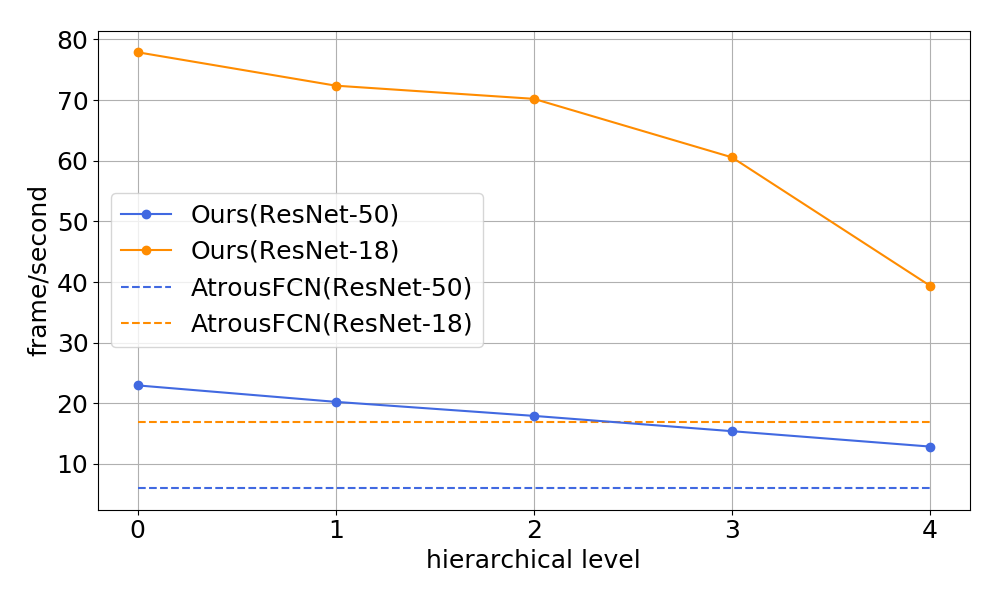}
    \caption{Inference speed for various hierarchical levels for ResNet-18 and ResNet-50 as the backbone. We report average fps over 100 trials for a 1024$\times$2048 input. Our proposed method is two times or more faster than AtrousFCN.}
    \label{fig:fps}
\end{figure}

Finally, we show frame per second (fps) in Fig. \ref{fig:fps}.
Our proposed method is two times or more faster than AtrousFCN.
Moreover, our method with the hierarchical level of two demonstrates higher mIoU than AtrousFCN, with sacrificing only 10\% fps compared with FCN-32.
Our method with ResNet-18 significantly decreases fps when the hierarchical level is four, because the model wastes the inference time for the clustering in the fine resolution feature map.

\subsection{Semantic Segmentation}
To show the effectiveness of our method against widely used models, we compare modern architectures, FCN-32~\cite{fcn}, FCN with atrous convolution (AtrousFCN)~\cite{atrous,dilated}, FPN~\cite{fpn}, PSPNet~\cite{pspnet}, and DeepLabv3~\cite{deeplabv3}, to the models adopting the proposed soft clustering on the Cityscapes~\cite{cityscapes} test set.
FCN-32 is the simplest model for dense prediction, and AtrousFCN is the detail-preserving model.
FPN is an asymmetric encoder-decoder model proposed for object detection, but it has also been used for a panoptic segmentation task~\cite{panofpn}.
PSPNet and DeepLabv3 are AtrousFCN combined with the spatial pyramid pooling.
The training protocol is the same as the ablation study.
We use ResNet-101~\cite{resnet} for the backbone architecture.
The ``HC'' models replace the backbone with our hierarchical clustering backbone illustrated in Fig. \ref{fig:overview}.
We set the hierarchical level of ``HC'' models to two.

\begin{figure*}[t]
        \centering
        \begin{minipage}{0.16\hsize}
            \centering
            \includegraphics[clip, width=1\hsize]{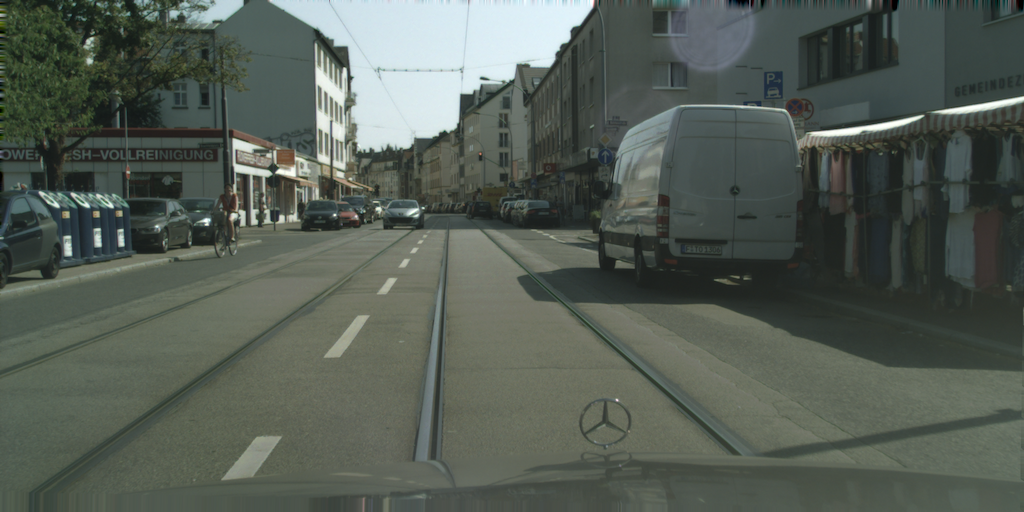}\\
            \includegraphics[clip, width=1\hsize]{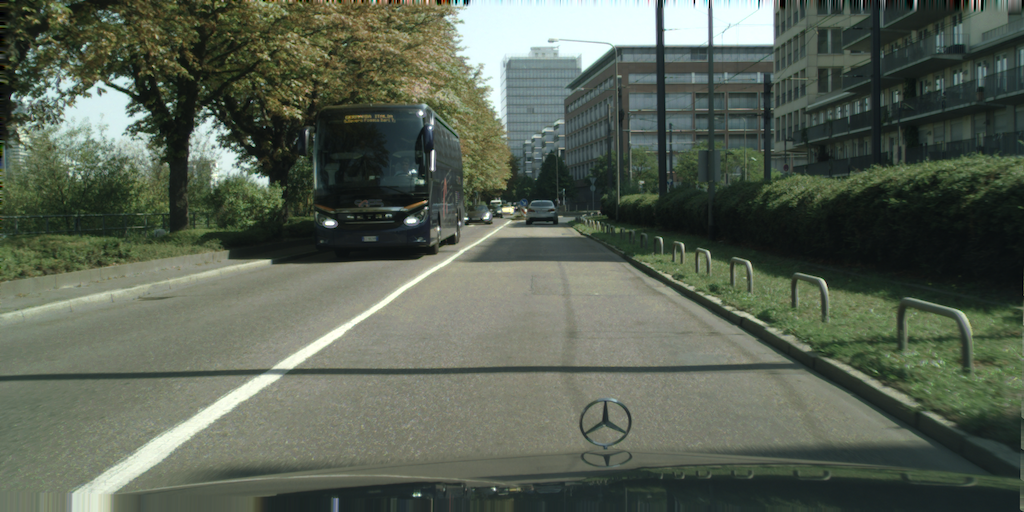}\\
            \includegraphics[clip, width=1\hsize]{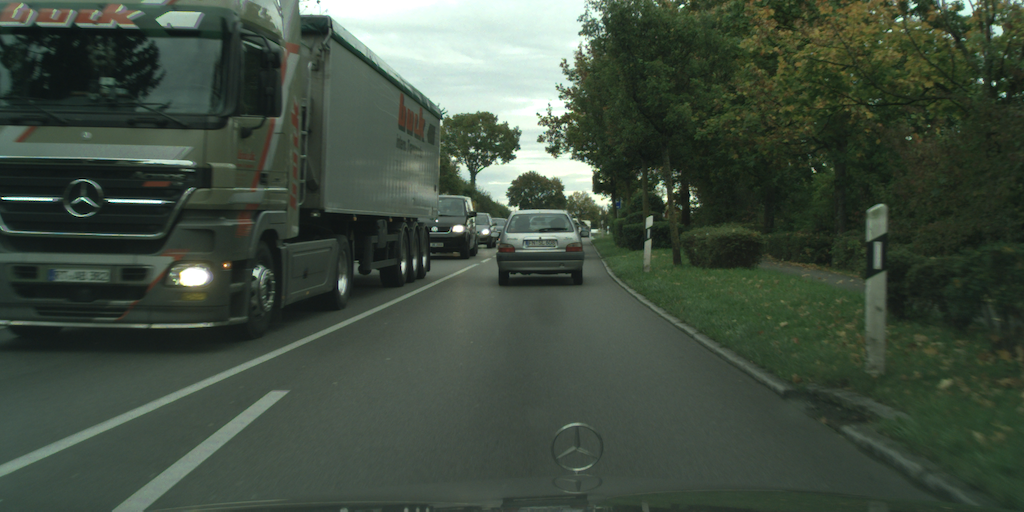}\\
            \includegraphics[clip, width=1\hsize]{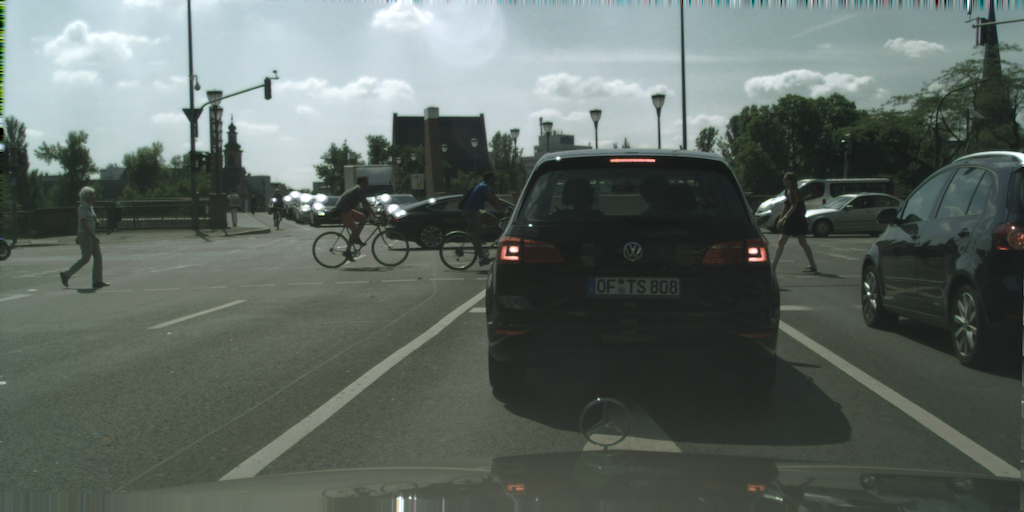}\\
            (a) Input
        \end{minipage}
        \begin{minipage}{0.16\hsize}
            \centering
            \includegraphics[clip, width=1\hsize]{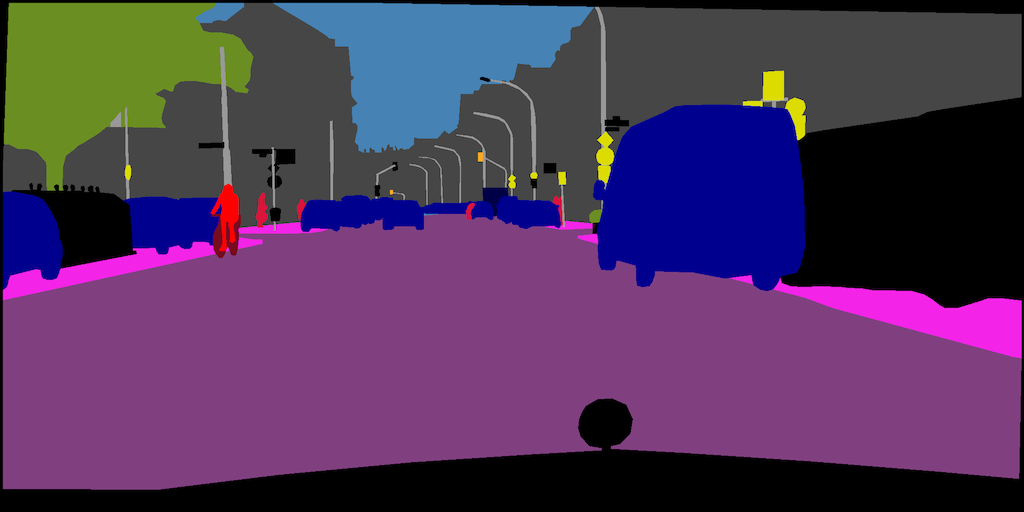}\\
            \includegraphics[clip, width=1\hsize]{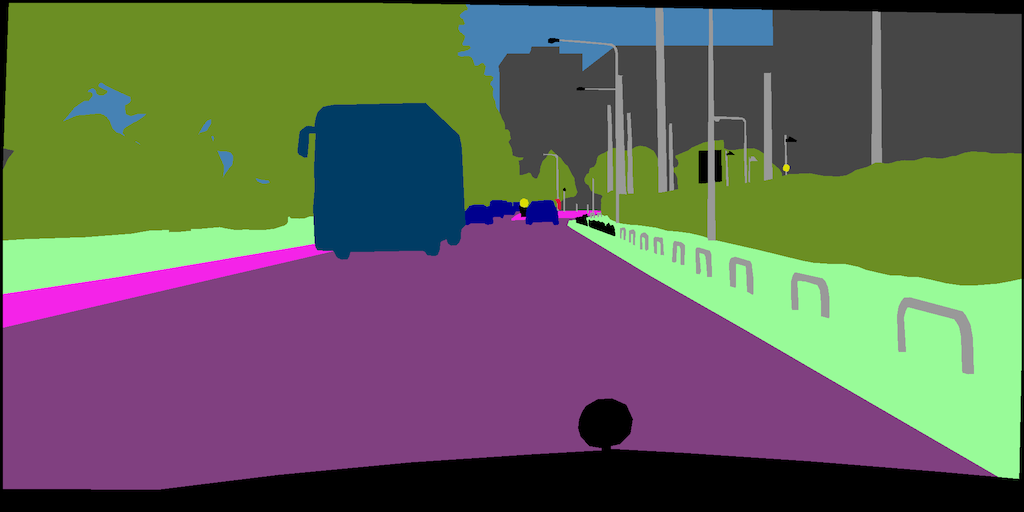}\\
            \includegraphics[clip, width=1\hsize]{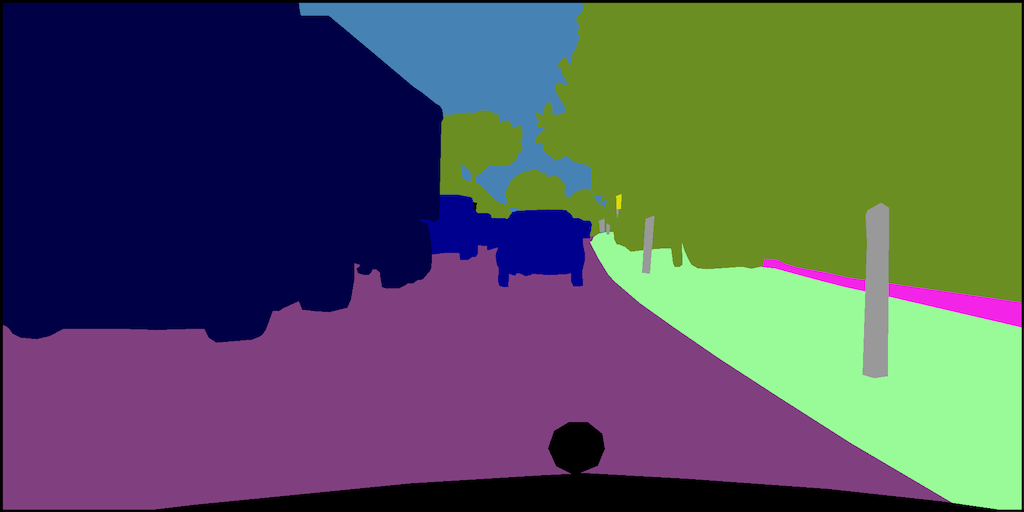}\\
            \includegraphics[clip, width=1\hsize]{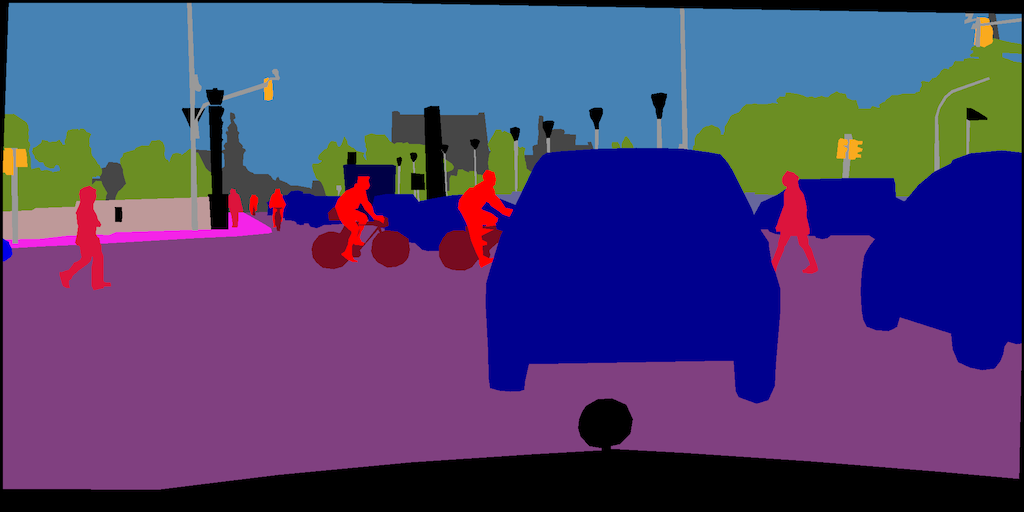}\\
            (b) Ground-truth
        \end{minipage}
        \begin{minipage}{0.16\hsize}
            \centering
            \includegraphics[clip, width=1\hsize]{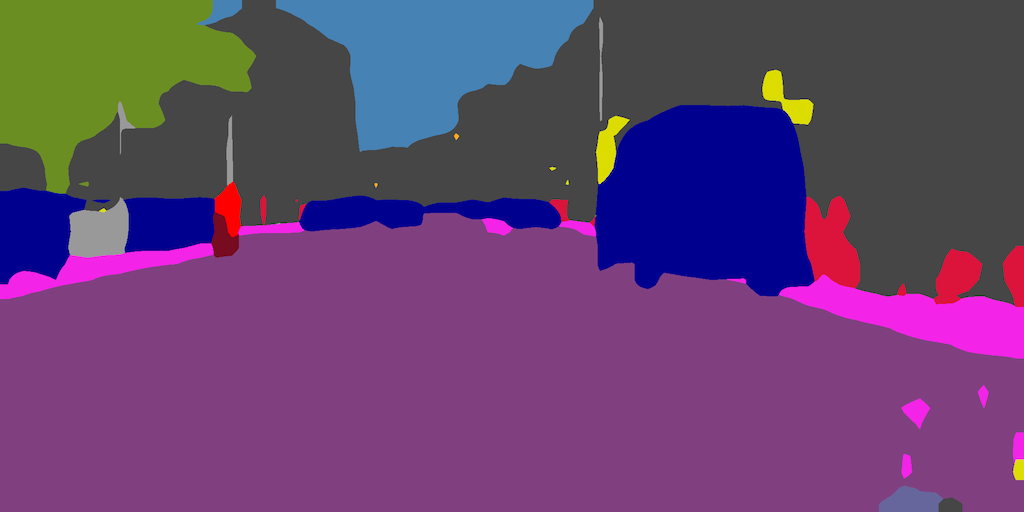}\\
            \includegraphics[clip, width=1\hsize]{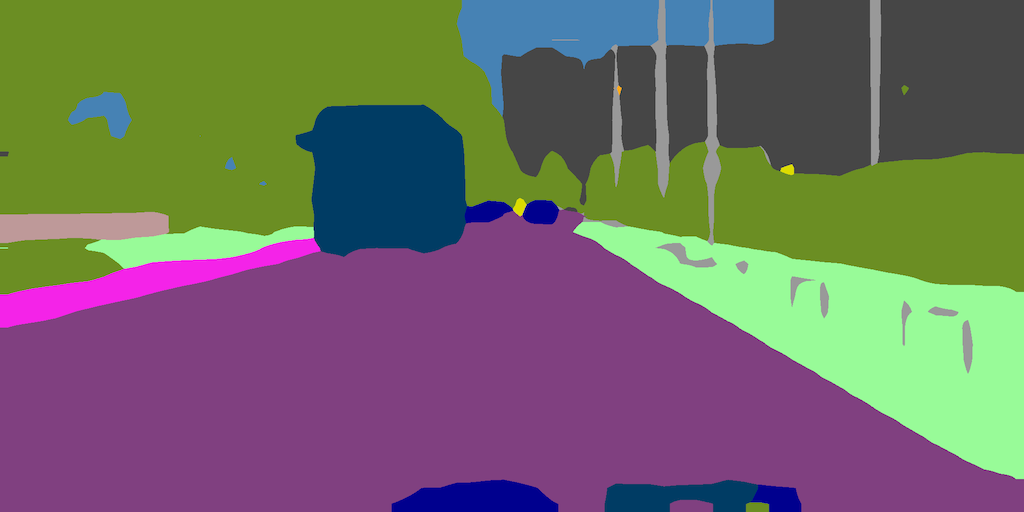}\\
            \includegraphics[clip, width=1\hsize]{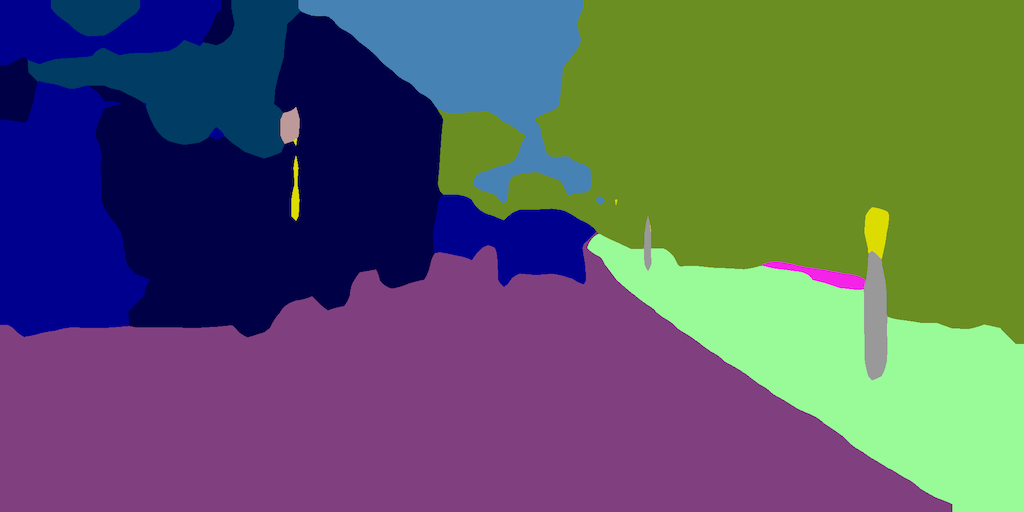}\\
            \includegraphics[clip, width=1\hsize]{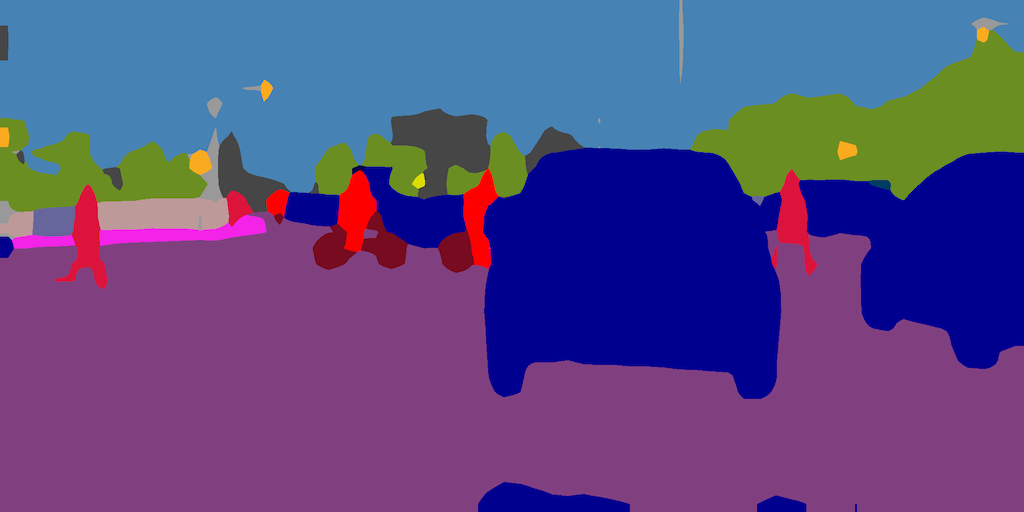}\\
            (c) FCN-32
        \end{minipage}
        \begin{minipage}{0.16\hsize}
            \centering
            \includegraphics[clip, width=1\hsize]{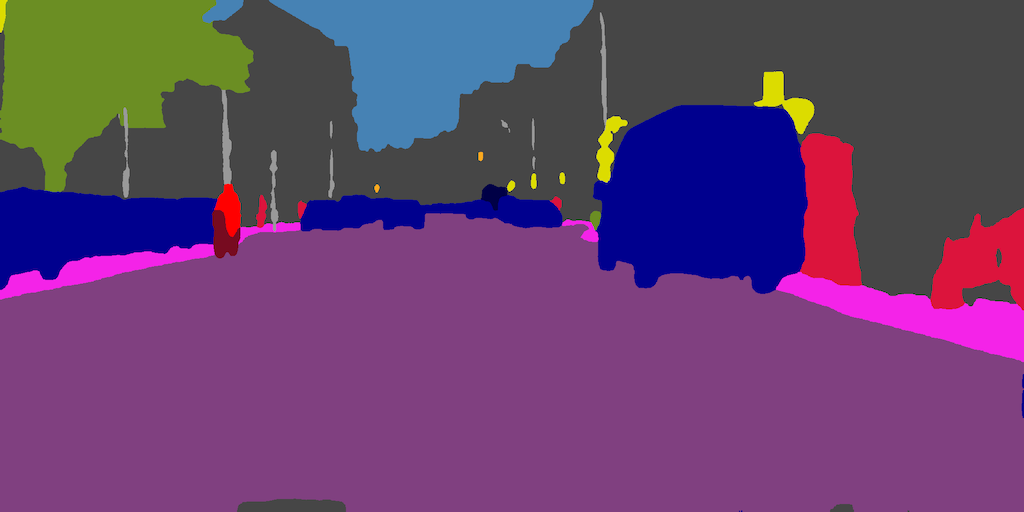}\\
            \includegraphics[clip, width=1\hsize]{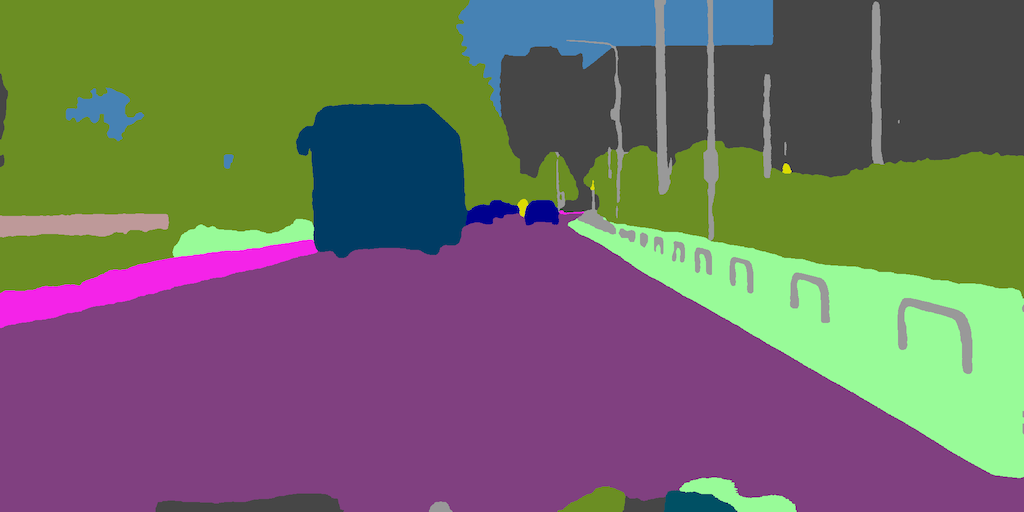}\\
            \includegraphics[clip, width=1\hsize]{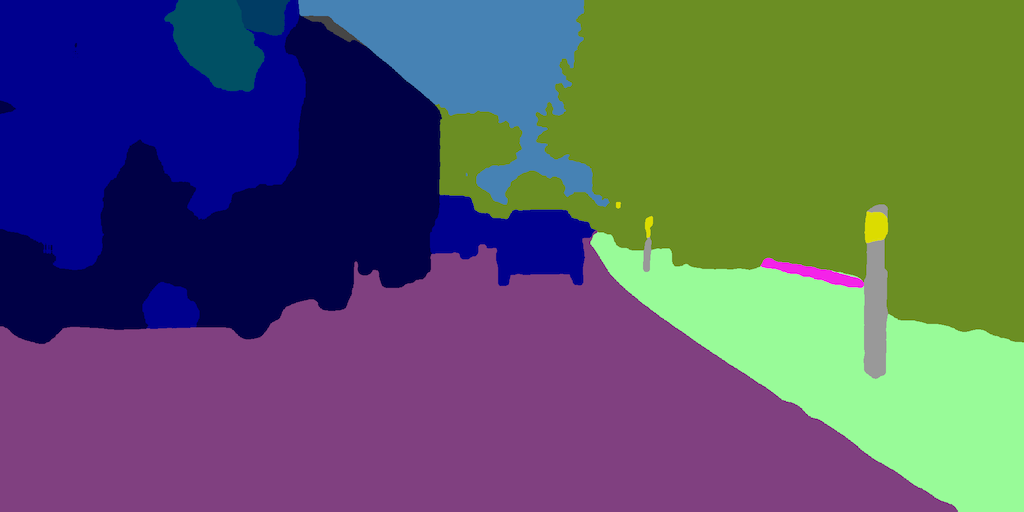}\\
            \includegraphics[clip, width=1\hsize]{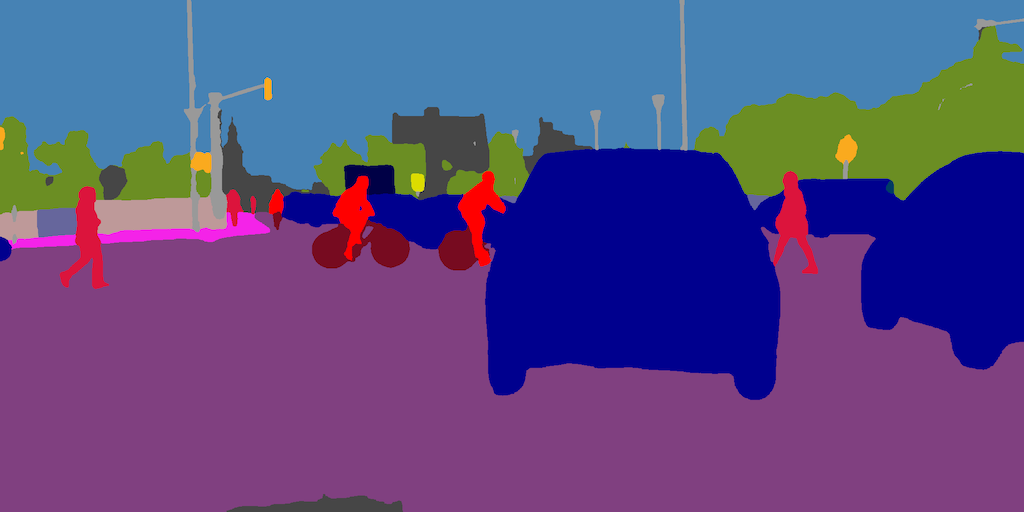}\\
            (d) HCFCN-32
        \end{minipage}
        \begin{minipage}{0.16\hsize}
            \centering
            \includegraphics[clip, width=1\hsize]{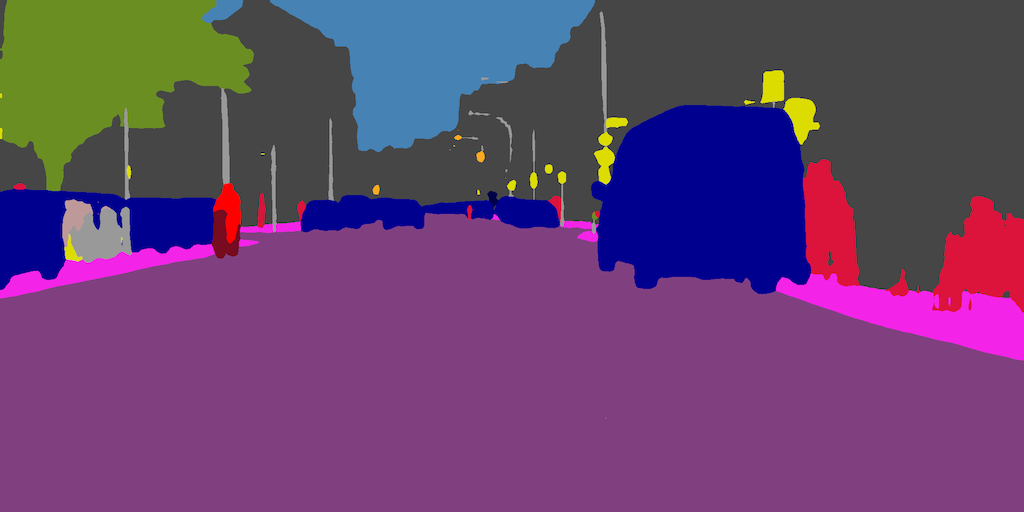}\\
            \includegraphics[clip, width=1\hsize]{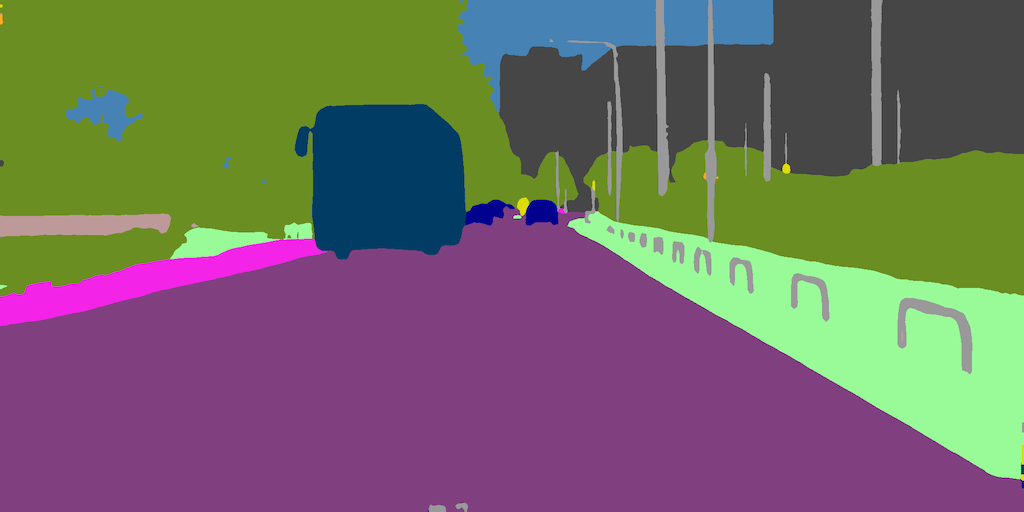}\\
            \includegraphics[clip, width=1\hsize]{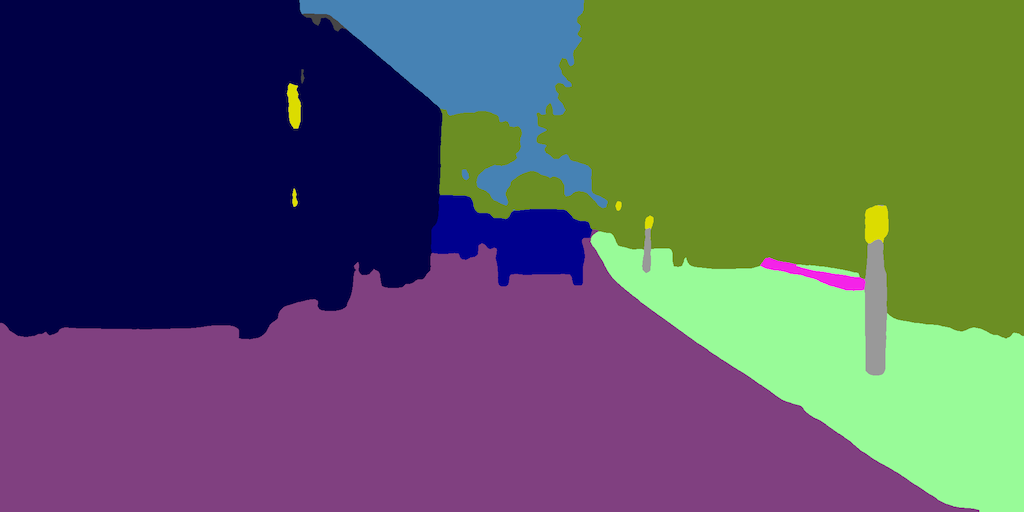}\\
            \includegraphics[clip, width=1\hsize]{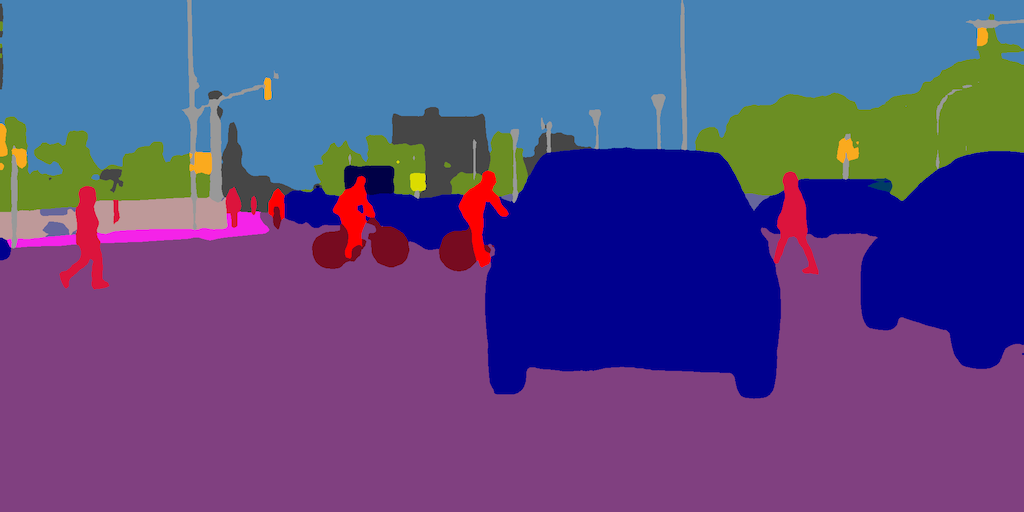}\\
            (e) PSPNet
        \end{minipage}
        \begin{minipage}{0.16\hsize}
            \centering
            \includegraphics[clip, width=1\hsize]{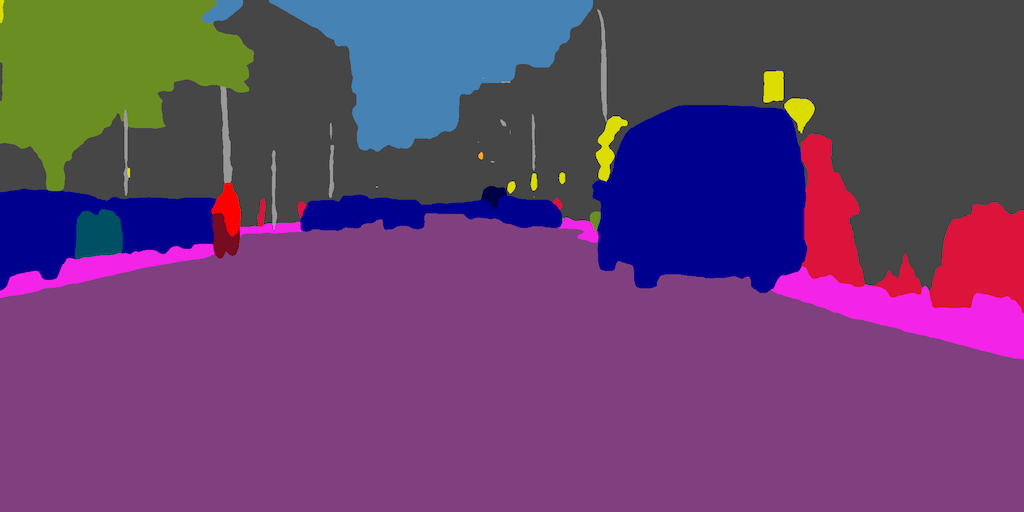}\\
            \includegraphics[clip, width=1\hsize]{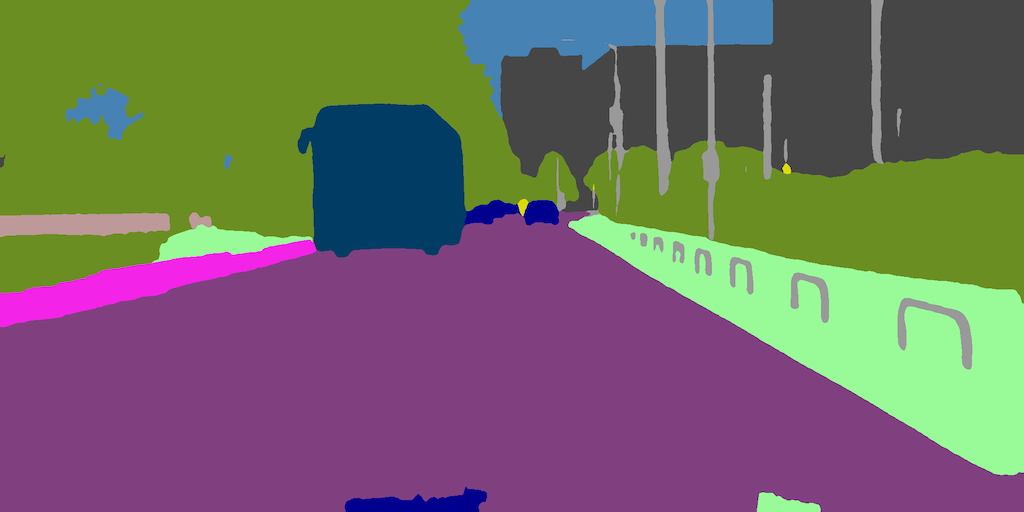}\\
            \includegraphics[clip, width=1\hsize]{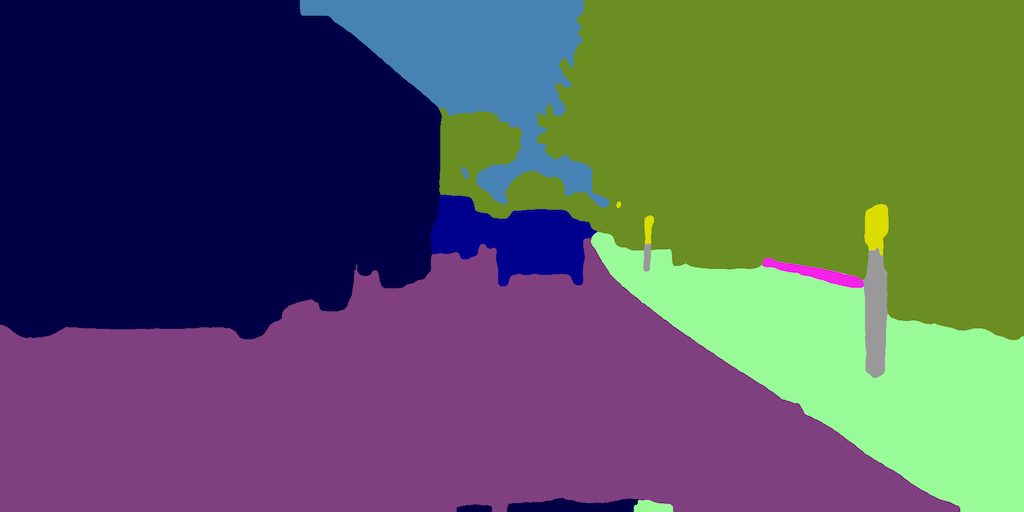}\\
            \includegraphics[clip, width=1\hsize]{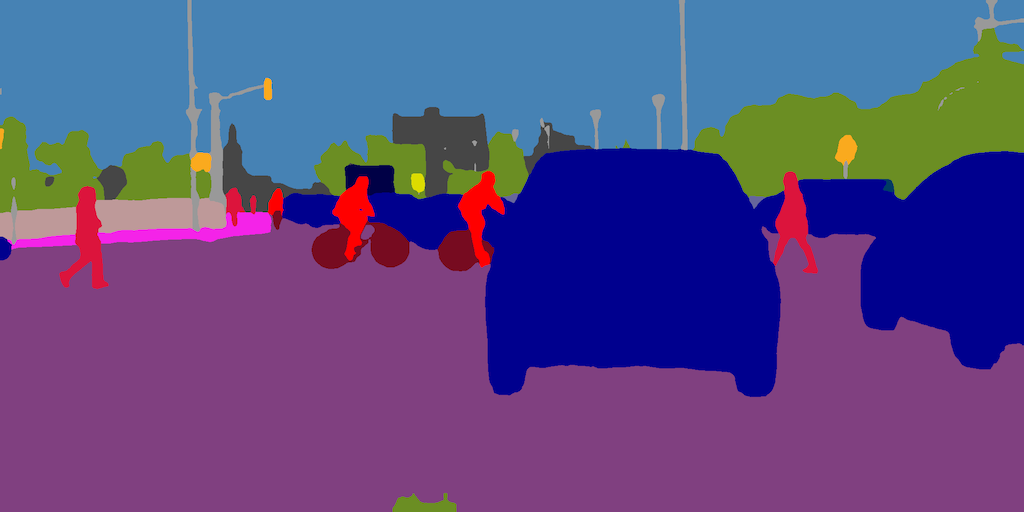}\\
            (f) HCPSPNet
        \end{minipage}
    \caption{Example results on the Cityscapes validation set. Each model uses ResNet-101 as the backbone. HCFCN-32 and HCPSPNet integrate the clustering modules into conv4\_x and conv5\_x in ResNet.}
    \label{fig:exm_seg}
\end{figure*}

We show mIoU in Tab. \ref{tab:city_res} and the example results in Fig. \ref{fig:exm_seg}.
Note that we evaluate the models with a multi-scale input and use the average results following~\cite{pspnet}.
\begin{table*}[t]
    \centering
    \scalebox{0.68}{
     \begin{tabular}{l|c|ccccccccccccccccccc||c}
        Model & OS & road & swalk & build. & wall & fence & pole & tlight & sign & veg. & terrain & sky & person & rider & car & truck & bus & train & mbike & bike & mIoU\\ \hline
        FCN-32  & 32 & 98.4 & 84.6 & 91.4 & 44.6 & 55.0 & 49.6 & 65.3 & 70.9 & 91.8 & 70.7 & 94.2 & 81.0 & 64.2 & 94.7 & 61.6 & 72.4 & 57.4 & 65.9 & 71.5 & 72.9 \\
        AtrousFCN & 8 & 98.5 & 85.6 & 92.6 & 47.1 & 54.9 & 66.9 & 74.1 & 77.8 & 93.2 & 72.0 & 95.5 & 85.7 & 70.5 & 95.4 & 58.7 & 72.0 & 61.4 & 69.3 & 75.4 & 76.1 \\
        FPN & 8 & 98.5 & 85.5 & 92.4 & 45.5 & 55.6 & 63.8 & 71.4 & 76.1 & 93.0 & 71.4 & 95.4 & 84.6 & 67.5 & 95.4 & 63.9 & 72.2 & 57.7 & 67.4 & 74.5 & 75.4\\
        PSPNet & 8 & 98.6 & 86.0 & 92.9 & 50.2 & 56.7 & 66.9 & 74.0 & 78.0 & 93.3 & 72.0 & 95.7 & 85.7 & 70.3 & 95.7 & 68.2 & 82.2 & 74.2 & 68.8 & 75.7 & 78.2\\
        DeepLabv3 & 8 & 98.6 & 86.5 & 93.0 & 48.5 & 57.1 & 67.5 & 74.4 & 78.4 & 93.3 & 72.5 & 95.6 & 86.0 & 72.0 & 95.7 & 71.4 & 82.4 & 74.3 & 69.7 & 75.8 & 78.6\\ \hline
        HCFCN-32 & 32 & 98.5 & 85.4 & 92.5 & 47.1 & 55.4 & 63.5 & 73.3 & 77.2 & 93.2 & 71.2 & 95.3 & 85.1 & 68.9 & 95.5 & 61.9 & 72.1 & 58.4 & 68.1 &75.1 & 75.7\\
        HCFPN & 8 & 98.5 & 85.2 & 92.5 & 47.4 & 55.4 & 64.7 & 73.4 & 77.1 & 93.2 & 71.2 & 95. 1 & 85.2 & 69.6 & 95.6 & 63.3 & 72.3 & 58.6 & 69.3 & 75.3 & 76.0\\
        HCPSPNet & 32 & 98.4 & 84.8 & 92.7 & 52.1 & 56.8 & 62.9 & 72.9 & 76.9 & 93.2 & 71.4 & 95.3 & 85.1 & 70.1 & 95.7 & 70.7 & 80.3 & 70.1 & 69.3 & 75.2 & 77.6\\
        HCDeepLabv3 & 32 & 98.6 & 85.7 & 92.7 & 52.2 & 55.6 & 62.8 & 73.0 & 76.9 & 93.2 & 71.7 & 95.4 & 85.0 & 69.8 & 95.6 & 69.2 & 81.3 & 72.8 & 68.2 & 74.7 & 77.6 \\
    \end{tabular}
    }
    \caption{Per-class results on the Cityscapes test set. We use ResNet-101 as the backbone and evaluate models with a multi-scale input following~\cite{pspnet}.}
    \label{tab:city_res}
\end{table*}
HCFCN-32 shows better mIoU than FCN-32.
Especially, HCFCN-32 improves mIoU for thin and small objects (\textit{e.g.}, pole, traffic light, and sign) that are often missed in the models that generate low-resolution maps.
In fact, HCFCN-32 predicts such objects better than FCN-32, as shown in Fig. \ref{fig:exm_seg}.
Our proposed method also improves mIoU for FPN, namely, the encoder-decoder model.
HCFPN also enhances 1\%-2\% mIoU for small objects from FPN.

However, HCPSPNet and HCDeepLabv3 show slightly lower mIoU than their original models.
Pyramid pooling modules used in them are proposed for atrous convolution based backbones, and their hyperparameters  (\textit{e.g.}, kernel sizes and pyramid scales) are likely tuned for high-resolution maps.
Since we adopt the same hyperparameters, they may not be the optimal values for low-resolution maps, namely, for our method.
% HCPSPNet and HCDeepLabv3 are a significant drop in mIoU for small and thin objects such as a pole and sign.
% We consider that the pyramid pooling module in PSPNet and DeepLabv3 impairs the detailed information for HCPSPNet and HCDeepLabv3 because their mIoU for small objects is lower than HCFCN-32.
% The pyramid pooling modules are useful to capture global information and enhance mIoU for large objects.
% In fact, PSPNet and DeepLabv3 significantly improve mIoU for the large objects compared with AtrousFCN.
% HCPSPNet and HCDeepLabv3 also significantly improve mIoU for the large objects but degrade about 1\% mIoU for small objects compared with HCFCN-32.
% Thus, the global context aggregation for low-resolution feature maps may impair the local context.

Atrous FCN shows slightly higher mIoU than HCFCN-32, although HCFCN-32 is better when the backbone is ResNet-18 and ResNet-50.
This fact may imply that our proposed method stably works, regardless of model size, but AtrousFCN with a shallow backbone is trapped in worse local minima.

We show the accuracy and inference time in Tab. \ref{tab:benchmark}.
\begin{table}[t]
    \centering
    \begin{tabular}{c|cc|c}
        Method & mIoU & Pixel acc. & msec/image \\ \hline
        FCN-32 & 71.7 & 94.9 & 71.7 \\
        AtrousFCN & 76.3 & 96.0 & 296.7 \\
        HCFCN-32 & 76.0 & 96.1 & 85.9 \\ \hline
        FPN & 75.1 & 95.8 & 82.0 \\
        HCFPN & 77.2 & 96.2 & 92.9 \\ \hline
        PSPNet & 77.7 & 96.2 & 298.7 \\
        HCPSPNet & 77.6 & 96.2 & 87.0 \\ \hline
        DeepLabv3 & 78.4 & 96.3 & 380.8 \\
        HCDeepLabv3 & 77.4 & 96.2 & 93.2
    \end{tabular}
    \caption{Accuracy and inference time of various architectures for a 1024$\times$2048 input on the Cityscapes validation set. The inference time is the average time over 100 trials. We evaluate the models with a single-scale input.}
    \label{tab:benchmark}
\end{table}
We report the average time over 100 trials on the NVIDIA Quadro RTX8000 GPU with a 1024$\times$2048 input and mIoU on the validation set.
Our proposed method significantly reduces latency for PSPNet and DeepLabv3 and improves mIoU for FPN and FCN-32 with a small increase in the inference time.

We also evaluate our proposed method on ADE20K~\cite{ade20k}.
ADE20K contains a broader range of scene and object categories than Cityscapes.
We train the models for 125k iterations with the same protocol as the ablation study except that we set the initial learning rate to 0.02.
As shown in Tab. \ref{tab:res_ade}, HCPSPNet and HCDeepLabv3 show comparable results to PSPNet and DeepLabv3, but our methods are significantly faster.

\subsection{Superpixel Segmentation}
We evaluate the soft clustering for superpixel segmentation on the BSDS500~\cite{bsds} test set.
BSDS500 contains 200 training images, 100 validation images, and 200 test images, and we use the training set for training.
We employ SSN~\cite{ssn} as a baseline method, a supervised superpixel segmentation method that consists of a shallow encoder-decoder model.
We introduce the soft clustering into SSN and replace bilinear upsampling operations with Eq. \eqref{eq:decode}.

We evaluate these models using achievable segmentation error (ASA) and boundary recall (BR).
ASA quantifies the achievable accuracy for given segmentation labels using superpixels as a pre-processing step, and BR assesses boundary adherence given ground-truth labels~\cite{spix_bench}.
\begin{table}[t]
    \centering
    \begin{tabular}{c|cc|c}
        Models & mIoU & Pixel Acc. & msec/image\\ \hline
        PSPNet & 42.53 & 80.91 & 48.74\\
        HCPSPNet & 42.56 & 80.85 & 18.83 \\ \hline
        DeepLabv3 & 42.70 & 81.04 & 62.51\\
        HCDeepLabv3 & 42.71 & 81.07 & 20.49\\
    \end{tabular}
    \caption{Results on the ADE20K validation set. The inference time is the average time over 100 trials on the NVIDIA Quadro RTX8000 with a 512$\times$512 input. We use ResNet-101 as the backborn architecture.}
    \label{tab:res_ade}
\end{table}

We set the initial learning rate to 5e-5 and decay it with cosine learning rate policy to 5e-7.
We employ Adam~\cite{adam} as optimizer and train models for 300K iterations with a batch size of 8.
We use random crop with a crop size of 208$\times$208 and random horizontal flip as the data augmentation.
The pixel feature dimension is set to 20.
The number of superpixels and differentiable SLIC iterations are set to 169 and 5 during training, respectively.
For testing, we set the iterations to 10.

\begin{figure}[t]
    \centering
    \includegraphics[clip,width=1\hsize]{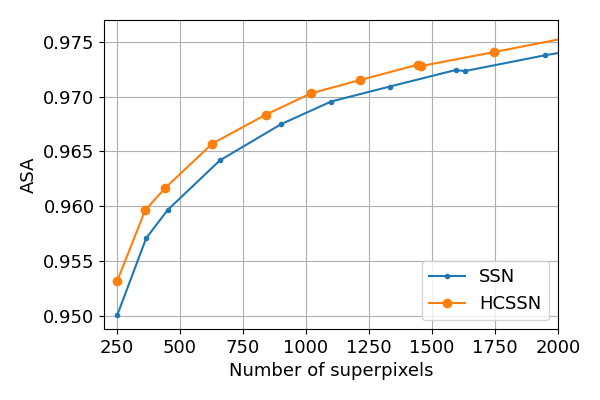}
    \caption{Achievable segmentation accuracy on BSDS500~\cite{bsds}.}
    \label{fig:asa}
\end{figure}

\begin{figure}[t]
    \centering
    \includegraphics[clip,width=1\hsize]{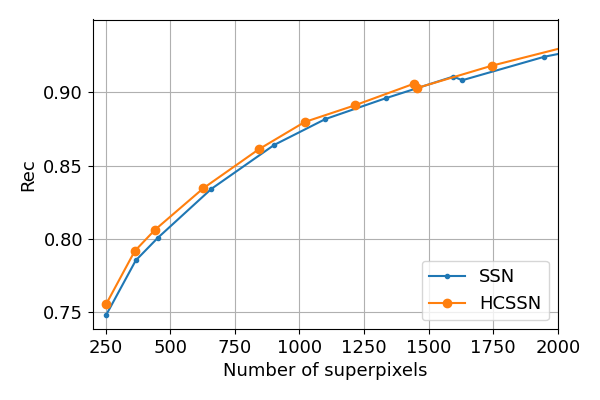}
    \caption{Boundary recall on BSDS500~\cite{bsds}.}
    \label{fig:br}
\end{figure}

\begin{figure}
    \centering
    \begin{tabular}{cc}
    \includegraphics[clip,width=0.5\hsize]{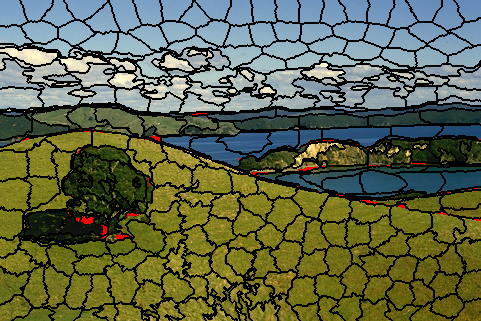}
    \includegraphics[clip,width=0.5\hsize]{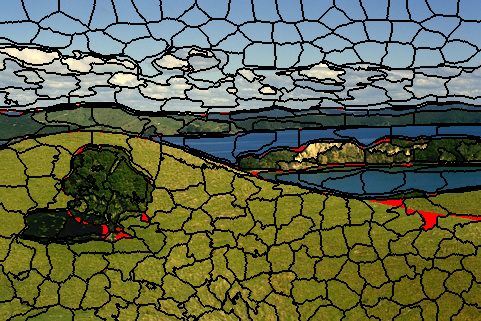}\\
    \includegraphics[clip,width=0.5\hsize]{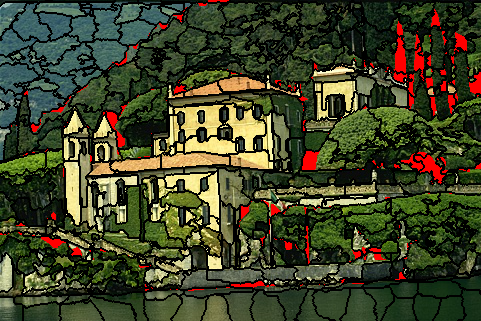}
    \includegraphics[clip,width=0.5\hsize]{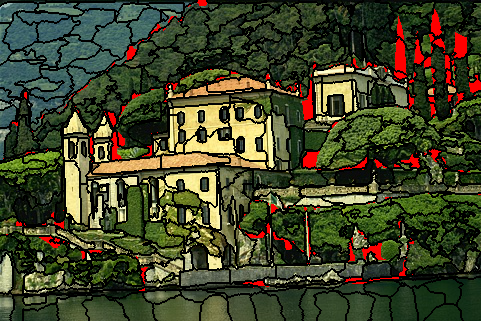}\\
    \includegraphics[clip,width=0.5\hsize]{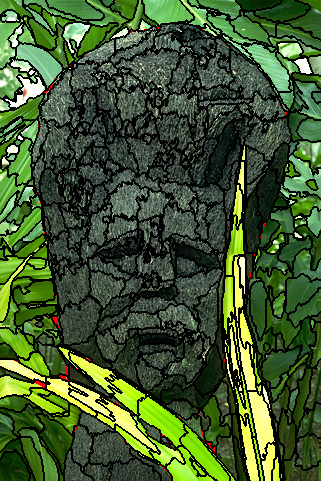}
    \includegraphics[clip,width=0.5\hsize]{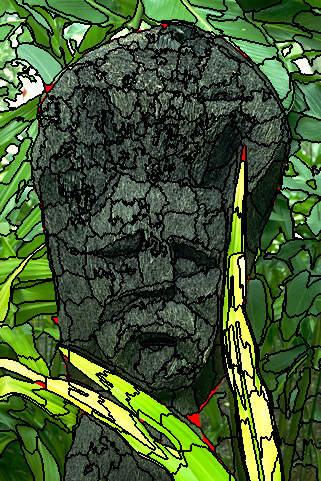}\\
    \includegraphics[clip,width=0.5\hsize]{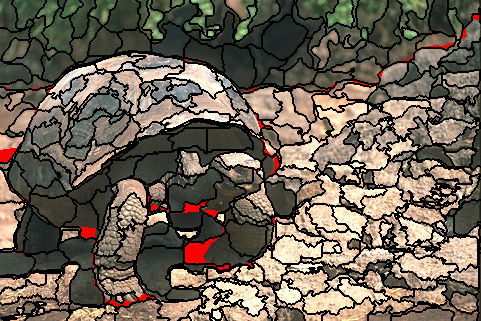}
    \includegraphics[clip,width=0.5\hsize]{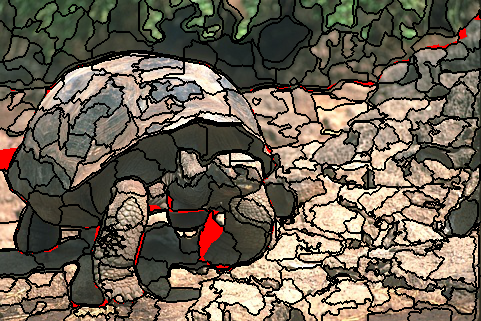}
    \end{tabular}
    \caption{Example results of HCSSN (left) and SSN (right). Red regions denote the undersegmentation error~\cite{spix_bench} that measures the ``leakage'' of superpixels with respect to ground-truth labels. Thus, images with less red areas are better.}
    \label{fig:spix_vis}
\end{figure}

We show ASA and BR in Figs. \ref{fig:asa} and \ref{fig:br}, and the example results in Fig. \ref{fig:spix_vis}.
The hierarchical soft clustering improves both ASA and BR, and as shown in Fig. \ref{fig:spix_vis}, we can observe that HCSSN reduces the undersegmentation error, which is a metric resembling $(1-\text{ASA})$~\cite{spix_bench}.

\subsection{Monocular Depth Estimation}
Our proposed method imposes local smoothness because of the clustering-based upsampling procedure, which may hurt the accuracy of the regression tasks.
Therefore, we also evaluate the proposed method for monocular depth estimation on NYU Depth v2~\cite{nyu}.

As a baseline method, we use DORN~\cite{dorn}, which consists of a dense feature extractor and a scene understanding module.
We compare DORN to a model that replaces the backbone in the dense feature extractor with our proposed backbone illustrated in Fig. \ref{fig:overview}.
Note that our model decodes a feature map generated by the scene understanding module with Eq. \eqref{eq:decode}, although the segmentation models decode the prediction map.
We train and evaluate the models with the same protocol as \cite{dorn}.
We report the average inference time over 100 trials on the NVIDIA Quadro RTX8000 with a 257$\times$353 input.


We show the comparison results in Tab. \ref{tab:depth_res}.
\begin{table}[t]
    \centering
    \scalebox{0.8}{
    \begin{tabular}{c|ccc|ccc|c}
        Model & $\delta_1$ & $\delta_2$ & $\delta_3$ & rel & $\log_{10}$ & rms & msec/image \\ \hline
        DORN~\cite{dorn} & 0.83 & 0.97 & 0.99 & 0.12 & 0.05 & 0.51 & 39.1\\
        HCDORN & 0.84 & 0.97 & 0.99 & 0.11 & 0.05 & 0.50 & 20.3
    \end{tabular}
    }
    \caption{Results on NYU Depth v2~\cite{nyu}.}
    \label{tab:depth_res}
\end{table}
Our soft clustering also produces a speedup without impairing the accuracy of the monocular depth estimation.

\section{Conclusion}
We proposed a way to implicitly integrate superpixel segmentation into existing FCN architectures without a change in their feed-forward path.
Our proposed method makes it easy to use superpixels with CNNs in an end-to-end manner.
% Although an ordinary downsampling layer misses the detailed information,
Our proposed method can preserve detailed information such as object boundaries by grouping pixels hierarchically in the downsampling layers.
As a result, our method improves mIoU and/or reduces inference time on the semantic segmentation task; for example, our proposed method enhances 2.8\% mIoU for FCN-32 and reduces inference time by 70\% for PSPNet on Cityscapes.
Moreover, our proposed method also demonstrates its effectiveness in other tasks; for example, our method produces a speedup or improves the accuracy of superpixel segmentation and monocular depth estimation tasks.

In this paper, we studied the effectiveness of our method against simple and widely used architectures.
We will investigate the effect of combining our method with more advanced methods in future work.
% In future work, we will investigate to combine our method with the advanced methods such as FAM and CARAFE.

{\small
\bibliographystyle{ieee_fullname}
\bibliography{egbib}
}

\newpage
\onecolumn
\appendix
\renewcommand{\thetable}{\Alph{section}.\arabic{figure}}
\renewcommand{\thefigure}{\Alph{section}.\arabic{table}}
\setcounter{figure}{1}
\setcounter{table}{1}

\section{DCNv2 as Downsampling Operation}
\label{sec:DCNvsSConv}
In our experiments, we use DCNv2~\cite{dconv2} as a downsampling operation because we consider that the flexibility of the downsampling operation is important to sample the effective cluster centers.
To verify it, we compare the strided convolution originally used in ResNet with DCNv2 with a stride of two under the same condition as the ablation study.
The training protocol is the same as the ablation study.
For FCN-32, we replace the strided convolutions except for \texttt{conv1} in ResNet with DCNv2 with a stride of two.

\begin{figure}[h]
    \centering
    \includegraphics[clip,width=0.7\hsize]{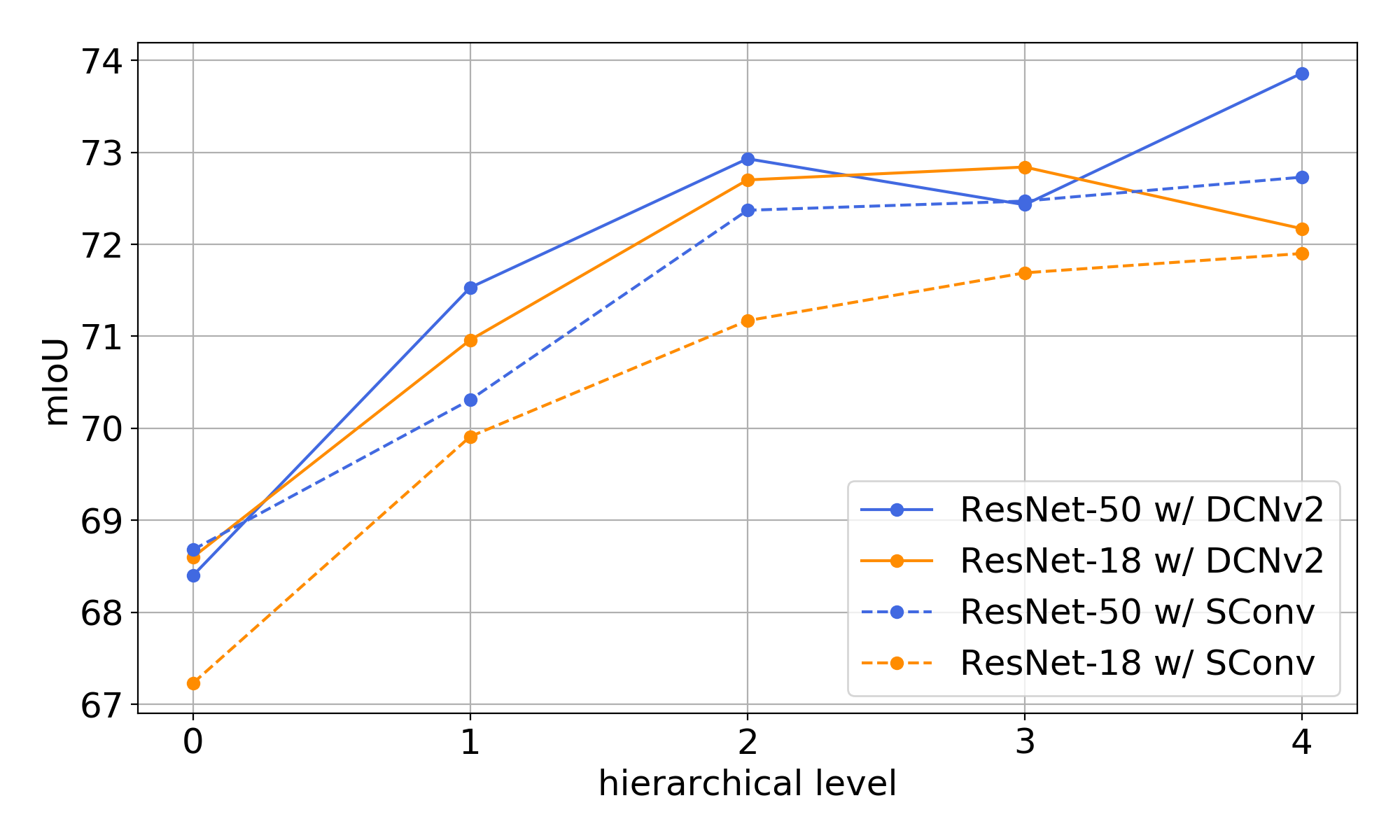}
    \caption{Comparison of the stride convolution (SConv) and DCNv2 as the downsampling layer.}
    \label{fig:sconv_vs_dconv}
\end{figure}

The results is shown in Fig. \ref{fig:sconv_vs_dconv}.
For FCN-32 with the ResNet-18 backbone, DCNv2 significantly improves mIoU compared with the strided convolution, but for ResNet-50, DCNv2 slightly decreases mIoU compared with the strided convolution.
For our method, DCNv2 stably improves mIoU.
Although DCNv2 is better than the strided convolution, our method with the strided convolution outperforms AtrousFCN when the hierarchical level is two or more.
AtrousFCN shows 70.90 and 72.06 mIoU on ResNet-18 and ResNet-50, and our method using the strided convolution shows 71.17 and 72.37 mIoU on ResNet-18 and ResNet-50 when the level is two.
Thus, DCNv2 is a better choice for our method, but the strided convolution also works effectively.

\section{Example PyTorch implementation of the proposed soft clustering}
\begin{lstlisting}[language=Python]
import torch
import torch.nn as nn
import torch.nn.functional as F


class SoftClustering(nn.Module):
    def __init__(self, feat_dim, downsampled_feat_dim, tau):
        super().__init__()
        self.W = nn.Conv2d(feat_dim, 64, kernel_size=1, bias=False)
        self.W_tilde = nn.Conv2d(downsampled_feat_dim, 64, kernel_size=1, bias=False)
        self.tau = tau

    def forward(self, features, downsampled_features):
        batch_size, n_channels, height, width = features.shape
        # map features into K-dimension space
        features = self.W(features)
        downsampled_features = self.W_tilde(downsampled_features)
        # get 9 candidate clusters and corresponding pixel features
        candidate_clusters = F.unfold(downsampled_features, kernel_size=3, padding=1).reshape(batch_size, n_channels, 9, -1)
        features = F.unfold(features, kernel_size=2, stride=2).reshape(batch_size, n_channels, 4, -1)
        # calculate similarities
        similarities = torch.einsum('bkcn,bkpn->bcpn', (candidate_clusters, features)).reshape(batch_size * 9, 4, -1)
        similarities = F.fold(similarities, (height, width), kernel_size=2, stride=2).reshape(batch_size, 9, height, width)
        # normalize
        soft_assignment = (similarities / self.tau).softmax(1)
        
        return soft_assignment

\end{lstlisting}

\section{Example PyTorch implementation of the cluster-based decoding (Eq. \eqref{eq:single_decode})}
\begin{lstlisting}[language=Python]
import torch
import torch.nn.functional as F


def decode(features, assignment):
    batch_size, _, height, width = assignment.shape
    n_channels = features.shape[1]
    # get 9 candidate clusters and corresponding assignments
    candidate_clusters = F.unfold(features, kernel_size=3, padding=1).reshape(batch_size, n_channels, 9, -1)
    assignment = F.unfold(assignment, kernel_size=2, stride=2).reshape(batch_size, 9, 4, -1)
    # linear decoding
    decoded_features = torch.einsum('bkcn,bcpn->bkpn', (candidate_clusters, assignment)).reshape(batch_size, n_channels * 4, -1)
    decoded_features = F.fold(decoded_features, (height, width), kernel_size=2, stride=2)

    return decoded_features
\end{lstlisting}

\end{document}